%% file: main.tex
\documentclass{article}

\usepackage[preprint]{neurips_2026}

\usepackage[utf8]{inputenc} %
\usepackage[T1]{fontenc}    %
\usepackage{url}            %
\usepackage{booktabs}       %
\usepackage{amsfonts}       %
\usepackage{nicefrac}       %
\usepackage{microtype}      %
\usepackage{xcolor}         %

\input{sections/preamble}

\title{Thinking Past the Answer: Evaluating Harmful Overthinking in Large Reasoning Models}

\author{%
  Simone Caldarella\textsuperscript{1,}\thanks{Correspondence to: \texttt{simone.caldarella@unitn.it}}\quad
  Davide Talon\textsuperscript{3}\quad \vspace{.1em}\\
  \textbf{Rahaf Aljundi\textsuperscript{2}}\quad
  \textbf{Elisa Ricci\textsuperscript{1,3}} \quad
  \textbf{Massimiliano Mancini\textsuperscript{1}}\vspace{1em} \\
  \textsuperscript{1}University of Trento \vspace{.1em}\\ \textsuperscript{2}Toyota Motor Europe \vspace{.1em}\\ \textsuperscript{3}Fondazione Bruno Kessler
}

\begin{document}

\maketitle

\vspace{-1em}
\input{sections/abstract}
\input{sections/introduction}
\input{sections/preliminaries}

\input{sections/method}

\input{sections/experiments}
\input{sections/related}
\input{sections/conclusion}

\acksection
The authors acknowledge the CINECA
award under the ISCRA initiative for the availability of high performance computing resources and support. This work
was supported by the EU Horizon ELIAS (No. 101120237),
ELLIOT (No. 101214398), and TURING (No. 101215032)
projects.

\medskip

{
\small
\bibliographystyle{plain}
\bibliography{bible}
}

\newpage
\appendix
\input{sections/supplementary}

\end{document}

%% file: sections/preamble.tex
\usepackage{multicol}
\usepackage{multirow}
\usepackage{tabularx}
\usepackage{xspace}
\usepackage{graphicx}
\usepackage{subcaption}
\usepackage{amsmath}
\usepackage{caption}
\usepackage{amssymb}
\usepackage[most]{tcolorbox}
\usepackage{inconsolata} %
\usepackage{tabularx}
\usepackage[most]{tcolorbox}
\usepackage{listings}
\usepackage{xcolor}
\usepackage{array}
\usepackage{booktabs}
\usepackage{xcolor}
\usepackage{algorithm}
\usepackage{enumitem}
\usepackage{pifont}

\usepackage{amsmath}
\usepackage{amsthm}
\usepackage{bm}
\usepackage{bbm}
\usepackage{xspace}
\usepackage{comment}
\usepackage{wrapfig}
\usepackage{algorithm}
\usepackage{algpseudocode}
\usepackage{graphicx}
\usepackage{subcaption}
\tcbuselibrary{listings, skins}
\usepackage{listings}
\usepackage{makecell}

\usepackage[table]{xcolor}
\usepackage[normalem]{ulem}

\definecolor{promptGray}{RGB}{247,247,250}
\tcbset{
  promptbase/.style={
    enhanced,
    colback=promptGray,
    left=10pt, right=10pt, top=6pt, bottom=8pt,
    arc=5pt,
    boxrule=0.8pt,
    attach boxed title to top left={xshift=12pt, yshift=-\tcboxedtitleheight/2},
    boxed title style={
      sharp corners, left=5pt, right=5pt, top=2pt, bottom=2pt, boxrule=0pt,
    },
    fonttitle=\small\bfseries\sffamily,
    fontupper=\ttfamily\small,
    coltitle=white,
  },
}

\definecolor{light}{rgb}{0.68, 0.90, 0.77}
\definecolor{orange}{rgb}{0.93, 0.74, 0.60}
\definecolor{lightorange}{rgb}{1, 0.87, 0.68}
\definecolor{lightgreen}{rgb}{0.76, 0.88, 0.76}
\definecolor{lightgray}{rgb}{0.95, 0.95, 0.97}
\definecolor{lightred}{rgb}{0.92, 0.29, 0.36}
\definecolor{multiflowcolor}{rgb}{0.92, 0.88, 1}
\definecolor{lightcyan}{rgb}{0.424, 0.651, 0.804}
\definecolor{zerocolor}{rgb}{0.91, 1, 0.999}

\definecolor{codeblue}{rgb}{0.25, 0.5, 0.5}
\definecolor{codekw}{rgb}{0.35, 0.35, 0.75}

\lstdefinestyle{Pytorch}{
    language         = Python,
    backgroundcolor  = \color{white},
    basicstyle = \fontsize{8.0pt}{9pt}\selectfont\ttfamily\bfseries,
    columns          = fullflexible,
    breaklines       = true,
    captionpos       = b,
    commentstyle     = \fontsize{4pt}{4pt}\color{codeblue},
    keywordstyle     = \fontsize{4pt}{4pt}\color{codekw},
    morekeywords     = {augment, softmax, confidence\_filter, torch, argmax},
}

\usepackage{multirow}

\definecolor{light}{HTML}{DBF2F4}

\definecolor{r1}{RGB}{73,129,226}
\definecolor{r2}{RGB}{218,29,103}
\definecolor{r3}{RGB}{221,167,36}
\definecolor{manuscript}{HTML}{4267B2}

\usepackage{pifont}

\newif\ifdraftmode
\draftmodetrue  %

\tcbuselibrary{listings,breakable,skins}

\newcommand{\ie}{\textit{i.e.}\xspace}
\newcommand{\eg}{\textit{e.g.}\xspace}

\tcbset{
  takeaway/.style={
    colback=gray!5,    %
    colframe=black!50, %
    boxrule=0.5pt,     %
    arc=2pt,           %
    left=4pt,right=3pt,top=3pt,bottom=3pt, %
    fonttitle=\bfseries,
  }
}

\definecolor{takeawayblue}{RGB}{22, 86, 160}
\definecolor{takeawaybg}{RGB}{245, 249, 255}
\definecolor{takeawayborder}{RGB}{150, 185, 225}

\newtcolorbox{takeawaybox}{
  enhanced,
  breakable,
  colback=takeawaybg,
  colframe=takeawaybg,
  boxrule=0pt,
  arc=3pt,
  left=9pt,
  right=8pt,
  top=3pt,
  bottom=3pt,
  borderline west={3pt}{0pt}{takeawayblue},
  before skip=0.5em,
  after skip=0.5em,
}

\newcommand{\range}[2]{#1,\dots, #2}

\renewcommand{\paragraph}[1]{\noindent\textbf{#1 }}

\newcommand{\vs}{\textit{vs.}}

\lstdefinestyle{promptstyle}{
  basicstyle=\ttfamily\small,
  columns=fullflexible,
  keepspaces=true,
  breaklines=true,
  breakatwhitespace=false,
  showstringspaces=false,
  frame=none,
  xleftmargin=0pt,
  xrightmargin=0pt
}

\tcbset{
  promptbase/.style={
    enhanced,
    colback=takeawaybg,
    colframe=takeawayborder,
    boxrule=0.6pt,
    arc=1mm,
    left=2mm,
    right=2mm,
    top=1.2mm,
    bottom=1.2mm,
    borderline west={2.2pt}{0pt}{takeawayblue},
    fonttitle=\bfseries,
    coltitle=takeawayblue,
    colbacktitle=takeawaybg,
    attach boxed title to top left={xshift=1.5mm, yshift*=-1mm},
    boxed title style={
      colback=takeawaybg,
      colframe=takeawaybg,
      boxrule=0pt
    }
  }
}

\usepackage{amsmath}
\DeclareMathOperator*{\argmin}{arg\,min}

\usepackage[
  colorlinks=true,
  citecolor=takeawayblue,
  linkcolor=takeawayblue,
  urlcolor=takeawayblue
]{hyperref}

%% file: sections/abstract.tex
\begin{abstract}
Large Reasoning Models (LRMs) improve performance by generating explicit intermediate reasoning traces through increased test-time compute, yet the assumption that longer reasoning is consistently beneficial remains under-examined.
While recent evidence shows that additional reasoning can lead models to overthink, we ask:
``\emph{Once a model has reached the correct answer, does further reasoning refine the solution, or deviate from it?}'' 
To study the dynamics after correctness, we introduce a prefix-level trajectory evaluation protocol grounded in reasoning sufficiency, defining the minimum reasoning budget required for a model to first generate the correct answer.
This allows us to disentangle \emph{verbose} overthinking, where additional reasoning is redundant but harmless, from \emph{harmful} overthinking, where continued reasoning destabilizes an already-correct trajectory. 
Starting from multimodal benchmarks, we find that many instances considered reasoning-intensive require surprisingly little reasoning. 
Moreover, stopping at the first correct prefix improves accuracy over standard reasoning up to 21\%, revealing that current models are limited not only by their ability to reason, but also by their inability to stop at the right time. 
Furthermore, while common efficiency strategies like early stopping substantially reduce verbose overthinking (up to 50\%), they fail to mitigate harmful overthinking. 
Failure analysis reveals that correctness deviations are mainly driven by logical drift and visual reinterpretation. 
Finally, we show that our findings generalize to language-only reasoning benchmarks, highlighting harmful overthinking as a broader reliability risk. 
Code available at \url{https://simonecaldarella.github.io/thinking-past-the-answer}.
\end{abstract}

%% file: sections/introduction.tex
\section{Introduction}

Large Reasoning Models (LRMs), such as OpenAI’s o1~\citep{jaech2024openai} and DeepSeek’s R1~\citep{guo2025deepseek}, have shown that allocating additional computation at test time can substantially improve performance on challenging tasks.\footnote{Throughout this paper, we use the term \emph{large reasoning models} to refer jointly to language-only and multimodal models trained to generate explicit intermediate reasoning traces.}
This paradigm, referred to as \emph{test-time scaling}~\citep{muennighoff2025s1}, improves performance by allowing models to produce longer and more deliberative reasoning traces, with gains observed in mathematical problems~\citep{hendrycks2021measuring,cobbe2021training}, code generation~\citep{chen2021evaluating,liu2023your}, and multimodal reasoning~\citep{lu2023mathvista,wang2024measuring}.
However, emerging evidence suggests that more reasoning is not always better: LRMs often exhibit systematic \emph{overthinking}, generating reasoning traces substantially longer than necessary to solve a problem~\citep{sui2025stop,liu2025efficient,chen2024not}.

Prior work has largely treated overthinking as an efficiency problem, aiming to reduce reasoning cost while preserving the accuracy of full-length chains of thought (CoT)~\citep{shen2025dast,zhang2025adaptthink,liu2025qfft,lin2025learning,xiao2025fastslow,wang2025make}.
This perspective has also been studied mostly in language-only settings~\citep{cuadron2025danger,wang2026thinking}, leaving limited insight into multimodal LRMs, where continued reasoning can introduce visual misreadings or unsupported reinterpretations of the input.
In this paper, we argue that this view is incomplete: overthinking is also a reliability problem.
A model may reach the correct answer early, continue reasoning, and later revise, contradict, or overwrite that correct solution.

\input{figures/teaser}

We study this phenomenon through the lens of \emph{reasoning sufficiency}.
For a given model and question, we define the question's difficulty as the minimum reasoning budget required for the model to produce the correct answer.
This differs from prior work that proxies difficulty using the average length of model-generated traces~\citep{sui2025stop,shen2025dast,lin2025learning}, since trace length can itself be inflated by overthinking.
Our formulation isolates the computation minimally required for correctness and separates two forms of overthinking:
\emph{verbose overthinking}, where the model reasons beyond the sufficient budget while preserving the correct answer, and
\emph{harmful overthinking}, where additional reasoning causes a trajectory that has already reached the correct answer to end with an incorrect final prediction.
Under this view, test-time scaling is not monotonically beneficial; additional computation can destabilize an already-correct solution.

To measure these effects, we introduce a \emph{prefix-level} trajectory evaluation protocol.
Given a reasoning trace, we evaluate prefix-level performance by forcing the model to produce an answer from that partial trace.
This lets us identify when the correct answer first becomes recoverable and whether continued reasoning preserves or loses correctness.
Using this protocol, we find that overthinking is substantial and systematic across multimodal benchmarks.
Many questions commonly viewed as reasoning-intensive can be solved with surprisingly few reasoning steps, yet models often continue far beyond the sufficient point.
As shown in Fig.~\ref{fig:teaser}, \emph{Optimal Length}, which stops at the first correct prefix, outperforms the model's default \emph{Actual Length} behavior by nearly \(10\%\) on average; this gain exceeds the benefit from reasoning-oriented post-training over the corresponding instruct model.
These results suggest that current LRMs are limited not only by whether they can reason, but also by whether they can stop reasoning at the right time.

We further show that harmful overthinking is not tied to a particular answer format or modality.
Both multiple-choice and free-form questions exhibit harmful overthinking, with surprisingly stronger effects in the latter settings, where the less constrained output space makes it easier to drift away from a previously correct answer.
Language-only experiments show that the same phenomenon also affects unimodal LRMs.
Moreover, simply shortening traces is insufficient: early stopping, \ie, terminating the reasoning trace earlier, reduces average reasoning length, but fails to mitigate harmful overthinking.
Finally, an analysis of \(4{,}842\) harmful traces reveals that correctness deviations are dominated by visual and logical errors, while calculation errors account for only a small fraction.

In summary, \textbf{the contributions of this paper} are as follows:
\begin{enumerate}[
  label=\ding{\numexpr171+\arabic*\relax},
  leftmargin=*,
  nosep
]
    \item We formalize overthinking via the minimum sufficient reasoning budget, disentangling verbose overthinking from harmful overthinking.
    \item We introduce a prefix-level evaluation protocol that measures reasoning sufficiency and correctness instability along model trajectories.
    \item We quantify harmful overthinking across multimodal and language-only benchmarks, showing that LRMs often drift from early correct answers to incorrect final predictions.
    \item We categorize the sources of harmful overthinking and show that correctness deviations are driven mainly by logical and visual errors rather than arithmetic mistakes.
\end{enumerate}

%% file: figures/teaser.tex
\begin{figure}[t]
    \centering
    \includegraphics[width=0.86\linewidth]{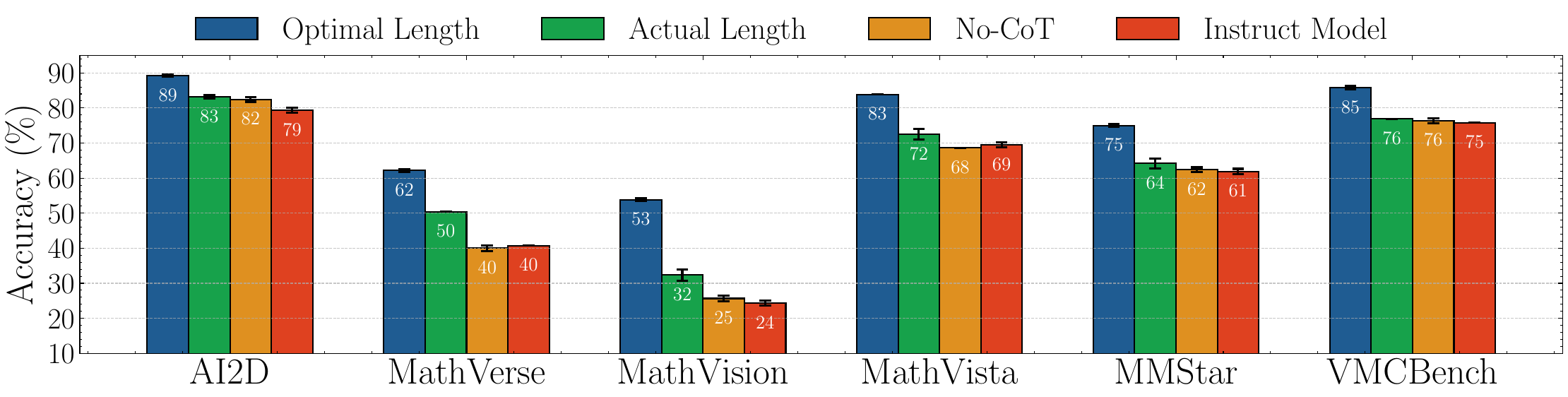}
    \caption{Performance averaged on LRMs. 
    \emph{Actual Length} is the model’s default behavior, \emph{No-CoT} disables intermediate reasoning, and \emph{Instruct Model} is the pre-reasoning instruction-tuned model.
    Finally, \emph{Optimal Length} stops at the first correct prefix.
    The gap between \emph{Actual Length} and \emph{Optimal Length} shows that models often reason past correctness, making additional reasoning harmful.}
    \label{fig:teaser}
    \vspace{-2em}
\end{figure}

%% file: sections/preliminaries.tex
\section{Formalizing Overthinking via Reasoning Sufficiency}
\label{sec:preliminaries}

In this section, we formalize overthinking through the lens of reasoning sufficiency.
We first define question difficulty as the minimum reasoning budget required for a model to reach a correct answer. 
We then use this notion to distinguish \emph{verbose} overthinking from \emph{harmful} overthinking.

\paragraph{Setting and Notation.} Let $(x,y)$ be a sample with input $x\in\mathcal X$ (potentially multimodal) and ground-truth answer $y\in\mathcal Y$. 
We consider a large reasoning model as a generative framework $\mathcal F:\mathcal X\rightarrow\mathcal T$ that, given $x$, produces a reasoning trace $t=\mathcal F(x)\in\mathcal T$ that includes the predicted answer, where $\mathcal T$ denotes the space of possible traces. 
For consistent evaluation in cases where answer formatting is not followed, we rely on a fixed answer extraction protocol where a language model $\mathcal A:\mathcal T\rightarrow\mathcal Y$ extracts the prediction from the provided trace $\hat y=\mathcal A(t)$. 
The extractor is implemented as a separate model (Qwen3-4B~\cite{yang2025qwen3}) that operates solely on the generated reasoning trace.

\subsection{Problem Difficulty}
\paragraph{\emph{What does it mean for a problem to be difficult?}}
Prior work often characterizes difficulty~\cite{shen2025dast,muennighoff2025s1,lin2025learning} using aggregate proxies such as \emph{pass@k} or average chain-of-thought length~\cite{shen2025dast}.
These proxies are confounded by decoding policy, sampling strategy, and verbosity, and therefore do not isolate the computation actually required for correctness.
We instead define the empirical difficulty of an instance \textit{wrt.} a model as the minimum reasoning budget (\ie, shortest CoT) sufficient for the model to obtain the correct answer.
This separates required reasoning from redundant/harmful continuation.

Formally, we consider $t$ as a sequence of $N$ utterances,
$t=(u_1,\dots,u_N)$,
where each $u_i$ represents a semantically coherent reasoning step. 
We denote by
$t_{\le i}=(u_1,\dots,u_i)$
the prefix up to step $i$, with $t_{\le 0}=\emptyset$ corresponding to no intermediate reasoning, \ie, the model can already answer without any reasoning.
Each prefix induces a prediction
$\hat y_i = \mathcal A(t_{\le i})$.
We define the \emph{first correct index} as:
\begin{equation}
    \tau_y(x;\mathcal F)
    =
    \argmin_{i\in\{0,\dots,N\}} b_i
    \quad \mathrm{s.t.} \quad
    \mathcal A(t_{\le i})=y,
\end{equation}
where $b_i$ is the computational budget associated with prefix $t_{\le i}, i=\range{0}{N}$,
and the empirical difficulty of the instance is $\hat\kappa(x,y;\mathcal F) = b_{\tau_y(x;\mathcal F)}$.
If no prefix yields the correct answer, we set $\tau_y=\infty$\footnote{Mathematicians may forgive us. In practice, when a trace does not reach the correct solution, we set the optimal length equal to the maximum length.} and leave $\hat\kappa$ undefined for that trajectory. 
We emphasize that $\hat{\kappa}(x,y;\mathcal F)$ is not an intrinsic property of the instance alone, but a \emph{model-dependent} difficulty of the sample.
This definition, invariant to overall length, captures the minimal computation required for the model to reach a correct answer: once a correct prefix has been reached, extending the reasoning does not change the difficulty. 
In practice, $\hat\kappa$ provides an \emph{empirical} lower bound on the compute required to form the correct answer.

\paragraph{On Tokens \vs  Utterances. } We instantiate the budget \(b_i\) as the number of utterances in \(t_{\le i}\). 
Unlike token count, utterance-level budgets are less sensitive to formatting and verbosity, and better align with semantically coherent reasoning steps.
In practice, we instantiate the reasoning steps by splitting traces at explicit delimiters (line breaks), which LRMs tend to use naturally. We use the generic $b_i$ to make clear that the definition is not tied to a particular notion of budget and the same definitions can be applied to token-level steps.
Appendix~\ref{supp:utterances_and_tokens} analyzes statistics on utterances and tokens.

\subsection{Disentangling Overthinking}
\paragraph{Verbose vs.\ Harmful.} Given the first correct index $\tau_y$, we define \emph{overthinking} as any continuation beyond the first correct prefix. 
That is, all steps $j>\tau_y$ correspond to computation that is not necessary to first obtain the correct answer. 
Then, by comparing the trace $t_{\leq \tau_y}$ with the full model one $t_{\leq N}$, we distinguish two cases:

\ding{172} \emph{Verbose} overthinking corresponds to wasted computation: once the model reaches a correct intermediate state, further reasoning does not change the outcome,
\begin{equation}
    \mathcal{A}(t_{\le \tau_y}) = y~~\land~~\mathcal{A}(t_{\le N}) = y.
\end{equation}
Here, additional reasoning is redundant. 
The model has already solved the problem, but continues to generate unnecessary steps without affecting the final prediction.

\ding{173} \emph{Harmful} overthinking, in contrast, reflects a failure of the reasoning process itself: after reaching a correct answer, additional computation causes the model to deviate from correctness,
\begin{equation}
    \mathcal{A}(t_{\le \tau_y}) = y~~\land~~\mathcal{A}(t_{\le N}) \neq y.
\end{equation}
In this case, the model initially reaches the correct solution, but subsequent reasoning introduces errors that override it, making the model reply incorrectly. 
Rather than refining the answer, additional computation destabilizes an otherwise correct trajectory.
Crucially, in Sec.~\ref{par:concise_model_harmful} we will show that reducing verbose overthinking does not reduce harmful overthinking, highlighting their orthogonality.

\paragraph{Harmful Overthinking as Trajectory Instability.}
The definition \ding{173} treats harmful overthinking as a binary event: after first reaching a correct answer, the model terminates with an incorrect one.
To analyze this behavior along the trajectory, we define the correctness state of each prefix as
\begin{equation}
    z_i = \mathbf{1}[\mathcal{A}(t_{\le i}) = y].
\end{equation}
Under monotonic reasoning, correctness would be absorbing: once \(z_i=1\), all later states would remain correct. 
Harmful overthinking corresponds to a violation of this monotonicity.

We therefore define the event-level harmful overthinking indicator as
\begin{equation}
    h(x;\mathcal{F})
    =
    \mathbbm{1}\left[\tau_y < \infty \ \wedge\ z_N = 0\right].
\end{equation}
Thus, \(h\) captures whether the model reaches a correct prefix but loses correctness by termination.
For a dataset \(\mathcal{D}\), we report the harmful overthinking rate as the average of this indicator:
\begin{equation}
    H(\mathcal{D};\mathcal{F})
    =
    \frac{1}{|\mathcal{D}|}
    \sum_{(x,y)\in\mathcal{D}} h(x;\mathcal{F}).
\end{equation}
Sec.~\ref{sec:overthinking} further analyzes reasoning  trajectory through the probability of remaining correct after \(\tau_y\).

%% file: sections/method.tex
\section{Overthinking in Large Reasoning Models}
\label{sec:overthinking}
We now define the main experimental protocol and investigate how reasoning unfolds in practice, relative to the minimum reasoning budget. Our analysis is guided by three questions: (i) how much reasoning is actually required to solve benchmark questions, (ii) what happens when models reason beyond this point, and (iii) whether reducing reasoning length mitigates potential failures.

We begin by examining how correct solutions first emerge along the reasoning trajectory, with a focus on the challenging multimodal setting.
Building on this perspective, we then study harmful overthinking, focusing on how additional reasoning can affect correctness. We further analyze how this phenomenon depends on the answer format, contrasting multiple-choice and free-form generation.
To better understand these effects, we adopt a prefix-level trajectory view and study correctness transitions across reasoning steps, revealing the underlying dynamics of reasoning. 
Finally, we evaluate whether reducing verbosity is sufficient to improve reliability, and assess the generality of these behaviors by extending the analysis to language-only models.

\subsection{Experimental Setting}

\paragraph{Models and Benchmarks.}
Building on prior work on overthinking~\cite{xiao2025fastslow,lin2025learning}, we analyze recent LRMs for multimodal reasoning: MM-Eureka~\cite{meng2025mm}, R1-VL~\cite{zhang2025learning}, ThinkLite-VL~\cite{wangsota}, and VL-Rethinker~\cite{vl-rethinker}.
We evaluate these models on a diverse set of multimodal benchmarks spanning diagram understanding, visual grounding, mathematical reasoning, and multiple-choice vision-language QA: AI2D~\citep{kembhavi2016diagram}, MathVista~\cite{lu2023mathvista}, MathVision~\cite{wang2024measuring}, MathVerse~\cite{zhang2024mathverse}, MMStar~\cite{chen2024we}, and VMCBench~\cite{zhang2025automated}. For language-only reasoning instead, we consider Qwen3~\cite{yang2025qwen3} and InternS1~\cite{bai2025intern} on  AIME2025~\cite{aime25} and GPQA~\cite{rein2023gpqa}.

\paragraph{Reasoning Strategies.} We evaluate four strategies spanning lower and upper bounds on reasoning performance. Instruct Model is the base \emph{Instruction-Tuned} model before reasoning-oriented post-training~\cite{zhang2026instruction}. \textit{No-CoT} forces the reasoning model to answer immediately, without intermediate reasoning. \textit{Actual Length} is the model's default unconstrained CoT behavior. \textit{Optimal Length} is an oracle strategy that stops at the first correct prefix $t_{\le \tau_y}$. Since identifying this prefix requires ground-truth access, it is not deployable; rather, it quantifies the gain achievable by eliminating harmful overthinking. 

\paragraph{Prefix-Level Trajectory Evaluation.} Inspired by previous work~\cite{fu2024efficiently, muennighoff2025s1}, we probe intermediate reasoning states by evaluating every utterance-level prefix $t_{\le i}$ of a generated trace, including the empty prefix $t_{\le 0}$. 
As reasoning models usually emit answers only at termination, we append a fixed termination template to each prefix\footnote{
``\texttt{Oh, I suddenly got the answer to the whole problem. <answer> \textbackslash n\textbackslash n \#\#\# Final Answer: [boxed\{}.''}, thus forcing the model to provide an answer at intermediate steps.
This intervention lets us track correctness across the reasoning trajectory by testing whether each \emph{partial} trace is sufficient to generate a correct answer. %
See Appendix~\ref{supp:early_termination_strategies} and~\ref{supp:prefix_level_evaluation} for more details.

\input{figures/optimal_vs_actual_length}

\subsection{Results}

\paragraph{How much reasoning is actually required?}
We first examine where the optimal stopping point occurs along the reasoning trajectory. Fig.~\ref{fig:optimal_vs_actual_length} compares, for solved instances, the model's actual reasoning length with the optimal length required to first reach the correct answer. Across benchmarks, optimal lengths are concentrated near the beginning of the trajectory, often at zero utterances, indicating the model can answer correctly without generating an explicit chain of thought. 
This is also confirmed by the performance that \emph{No-CoT} achieves across benchmarks (see Fig.~\ref{fig:teaser}).
On the contrary, actual traces extend substantially further.
Notably, even on more challenging datasets such as MathVision and MathVerse, where traces are longer than on AI2D or VMCBench, the optimal length remains far below the model's default reasoning length. 

\begin{takeawaybox}
\textsc{Takeaway A.} Reasoning length is a poor proxy for difficulty: LRMs often solve the problem early, then keep generating long traces that are not required for correctness.
\end{takeawaybox}

\paragraph{Reasoning beyond optimal.}
\input{tables/difficulty_len_malign}
We next quantify the effect of reasoning beyond the first correct step $t_{\leq\tau_y}$. From Tab.~\ref{tab:main_tab}, \emph{Optimal Length} consistently outperforms \emph{Actual Length} across all models and benchmarks (\eg, +23.3\% of R1-VL on MathVision and +7.8\% of VL-Rethinker on AI2D). 
The largest gaps occur on harder, lower-accuracy benchmarks such as MathVision and MathVerse.
This gap is not merely an efficiency loss.
In many cases, the model has already reached the correct answer, but later reasoning causes it to deviate from the correct answer. 
Together with Fig.~\ref{fig:teaser}, these results show that allocating the right amount of reasoning is often more important than simply enabling reasoning: the gap between \emph{Optimal Length} and \emph{Actual Length} is larger than the gain from reasoning-oriented post-training itself. 
See Appendix~\ref{supp:verbose-overhead} for analysis of verbose overthinking.

\begin{takeawaybox}
\textsc{Takeaway B.} Current LRMs do not merely over-generate reasoning; instead, they frequently reason past correct intermediate states, making optimal stopping substantially more valuable than additional reasoning.
\end{takeawaybox}

\paragraph{Multiple-choice \textit{vs.} Free-form.}
We next ask whether harmful overthinking depends on the answer format. 
Fig.~\ref{fig:mc_ff_overthinking} compares multiple-choice (MC) and free-form (FF) questions, aggregated across benchmarks. Harmful overthinking is substantially higher in free-form  (.11 for MC~\vs{} .24 for FF), suggesting that earlier correctness and later deviations are not byproducts of a restricted answer space, but rather the opposite.
If correctness deviations were primarily random answer fluctuations, one would expect multiple-choice tasks to exhibit higher, or at least comparable, earlier correctness and later answer instability. 
Surprisingly, we observe the opposite pattern.
This suggest that (i) earlier correct answers are not byproduct of randomness and (ii) correctness is less stable when the setting involves verification (MC) rather than exploration (FF),
making correctness in FF setting more vulnerable to unsupported revisions, reinterpretations, and reasoning drift.

\begin{takeawaybox}
\textsc{Takeaway C.} Free-form generation exposes harmful overthinking more sharply: without a fixed answer set, the unconstrained reasoning is more likely to deviate from correctness.
\end{takeawaybox}

\paragraph{Reasoning Dynamics.}
The previous results measure whether harmful overthinking occurs.
We now examine how it occurs by tracking correctness along the reasoning trajectory. At each prefix $t_{\le i}$, the model is either correct or incorrect, inducing a binary state $z_i=\mathbf{1}[\mathcal{A}(t_{\le i})=y]$. If reasoning were monotonic, then reaching a correct state would be absorbing: once $z_i=1$, subsequent prefixes would remain correct.
Instead, Fig.~\ref{fig:stability_survival} shows that correctness is unstable under continued generation. 
After the first correct prefix $\tau_y$, the probability of remaining correct drops rapidly as additional reasoning steps are generated, plateauing around 0.2 after roughly 100 intermediate steps.
This confirms the trajectory-instability view: reasoning does not simply accumulate evidence toward correctness, but can move the model both toward and away from the correct answer.

\begin{takeawaybox}
\textsc{Takeaway D.} Reasoning trajectories are non-monotonic: after reaching correctness, the probability of staying correct drops rapidly as LRMs keep reasoning.
\end{takeawaybox}

\input{tables/early_stopping_and_efficiency}

\input{figures/stability_and_multichoice}

\label{par:concise_model_harmful}

\paragraph{Can reducing verbosity mitigate harmful overthinking?}
A natural hypothesis is that harmful overthinking is simply a consequence of verbosity: if a model reasons less, it should have fewer opportunities to leave a correct trajectory. We test this hypothesis with two forms of adaptive inference.
First, we consider a training-free early-stopping baseline, applied to each model, inspired by prior work~\cite{fu2024efficiently}: after each prefix, we extract the current answer and stop when the prediction remains unchanged for $K$ consecutive steps (\ie, Stopping@K).
The setting $K=\infty$ recovers the model's default behavior, \ie, \emph{Actual Length}. Second, we compare VL-Rethinker~\cite{vl-rethinker}, trained to encourage thinking, against DualMind-VLM~\cite{lin2025learning}, which is explicitly trained to select whether to use reasoning or not.

Tab.~\ref{tab:early_stopping_efficiency} shows that both approaches substantially reduce reasoning length. Early stopping with smaller patience values cuts the average length from $17.2$ utterances at $K=\infty$ to $8.9$ for $K=5$ and $5.1$ for $K=2$. Similarly, DualMind-VLM produces much shorter traces than VL-Rethinker ($11.2$ vs.\ $23.4$ utterances) while maintaining comparable accuracy. However, this reduction in verbosity does not translate into a corresponding reduction in harmful overthinking. In fact, early stopping increases the harmful overthinking rate from $10.1$ at $K=\infty$ to $18.4$ at $K=5$ and $14.1$ at $K=2$, while DualMind-VLM still exhibits non-negligible harmful transitions despite its shorter traces.
These results show that verbose and harmful overthinking are distinct failure modes. Reducing the amount of generated reasoning can remove wasted computation, but it does not necessarily make the remaining trajectory more stable. In some cases, aggressive stopping may even truncate useful recovery dynamics while leaving correctness deviations unresolved.

\begin{takeawaybox}
\textsc{Takeaway E.} Efficiency-oriented methods address verbose reasoning, but not correctness instability. 
\textit{Harmful} overthinking must therefore be measured separately from \emph{verbose} one.
\end{takeawaybox}

\input{tables/difficulty_len_malign_language}

\paragraph{Language-Only Reasoning.}
Finally, we verify that the pattern is not specific to multimodal reasoning. Tab.~\ref{tab:main_tab_language} shows the same qualitative behavior for language-only LRMs on GPQA and AIME2025: default reasoning improves over \emph{No-CoT}, but \emph{Optimal Length} yields much larger gains. For Qwen3, optimal stopping improves accuracy from $55.8$ to $77.9$ on GPQA and from $58.3$ to $91.7$ on AIME2025. Similarly, InternS1 improves from $64.4$ to $84.7$ on GPQA and from $38.9$ to $72.2$ on AIME2025. 
These gains coincide with large reductions in reasoning length. 
For example, Qwen3 on AIME2025 drops from $372.5$ to $29.9$ utterances under \emph{Optimal Length}. 
Thus, our experiments show that harmful overthinking is not an artifact of visual grounding but reflects a broader instability of the reasoning process. 
See Appendix~\ref{supp:language_reasoning} for more results on the language-only setup.

\begin{takeawaybox}
\textsc{Takeaway F.} 
Harmful overthinking is not merely a byproduct of visual drift or instability in multimodal reasoning: similar patterns also appear in language-only models, even on math-heavy complex benchmarks.
\end{takeawaybox}

%% file: figures/optimal_vs_actual_length.tex
\begin{figure}
    \centering
    \includegraphics[width=0.8\linewidth]{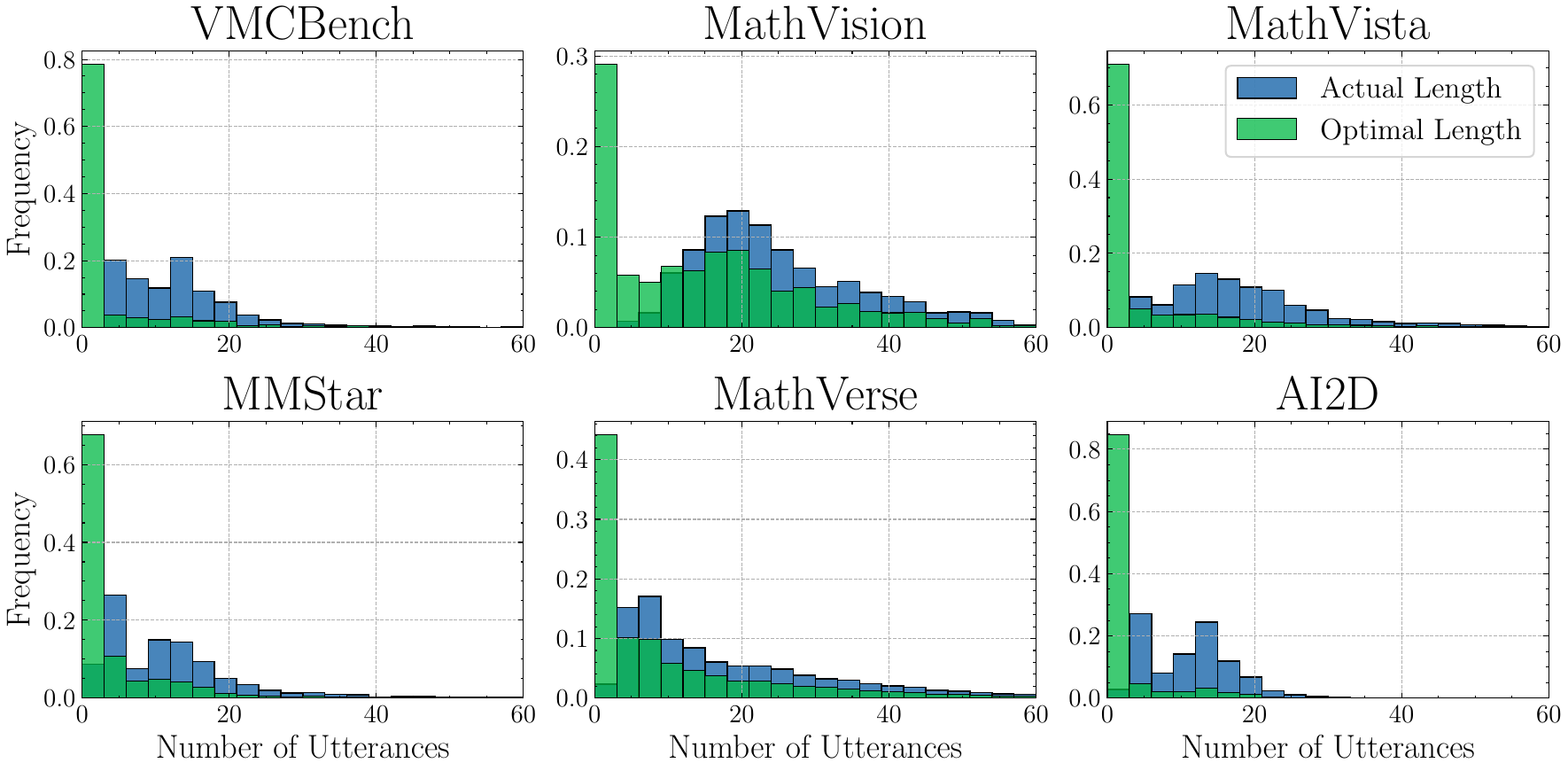}
    \caption{Average number of utterances across five multimodal models under Actual Length and Optimal Length. Even on benchmarks typically considered challenging (\eg, Mathvision~\cite{ wang2024measuring})
    most solvable instances require little to no intermediate reasoning. %
    }
    \label{fig:optimal_vs_actual_length}
\end{figure}

%% file: tables/difficulty_len_malign.tex
\begin{table}[t]
    \centering
    \scriptsize
    \caption{
Main multimodal results. We report accuracy (\(\mathrm{acc}\uparrow\)), average utterance length (\(\mathrm{len}\downarrow\)), and harmful-overthinking rate (\(H\downarrow\)).
\emph{No-CoT} is a zero-reasoning diagnostic; Bolding highlights the best nontrivial reasoning strategy.
The gap between \emph{Actual} and \emph{Optimal} shows that LRMs often reason past correctness and degrade final performance.
}
\vspace{1em}
    \label{tab:main_tab}
    \renewcommand{\arraystretch}{1}
    \begin{tabularx}{\linewidth}{
        >{\raggedright\arraybackslash}p{45pt}
        >{\centering\arraybackslash}p{24pt}
        >{\centering\arraybackslash}X 
        >{\centering\arraybackslash}X
        >{\centering\arraybackslash}X @{\hspace{16pt}}
        >{\centering\arraybackslash}X 
        >{\centering\arraybackslash}X
        >{\centering\arraybackslash}X @{\hspace{16pt}}
        >{\centering\arraybackslash}X 
        >{\centering\arraybackslash}X
        >{\centering\arraybackslash}X @{\hspace{16pt}}
        >{\centering\arraybackslash}X 
        >{\centering\arraybackslash}X
        >{\centering\arraybackslash}X @{\hspace{16pt}}
        >{\centering\arraybackslash}X 
        >{\centering\arraybackslash}X
        >{\centering\arraybackslash}X @{\hspace{16pt}}
        >{\centering\arraybackslash}X 
        >{\centering\arraybackslash}X
        >{\centering\arraybackslash}X
    }
    \toprule
    \multirow{2.5}{*}{\textbf{Model}} & \multirow{2.5}{*}{\textbf{Strategy}} & \multicolumn{3}{c}{\textbf{VMCBench}} & \multicolumn{3}{c}{\textbf{MathVision}} & \multicolumn{3}{c}{\textbf{Mathvista}} & \multicolumn{3}{c}{\textbf{MMStar}} & \multicolumn{3}{c}{\textbf{MathVerse}} & \multicolumn{3}{c}{\textbf{AI2D}} \\
    \cmidrule(lr){3-5} \cmidrule(lr){6-8} \cmidrule(lr){9-11} \cmidrule(lr){12-14} \cmidrule(lr){15-17} \cmidrule(lr){18-20}
    & & \textit{acc$\uparrow$} & \textit{len$\downarrow$} & \textit{$H$$\downarrow$} & \textit{acc$\uparrow$} & \textit{len$\downarrow$} & \textit{$H$$\downarrow$} & \textit{acc$\uparrow$} & \textit{len$\downarrow$} & \textit{$H$$\downarrow$} & \textit{acc$\uparrow$} & \textit{len$\downarrow$} & \textit{$H$$\downarrow$} & \textit{acc$\uparrow$} & \textit{len$\downarrow$} & \textit{$H$$\downarrow$} & \textit{acc$\uparrow$} & \textit{len$\downarrow$} & \textit{$H$$\downarrow$}\\
\midrule
\multirow{4}{*}{DualMind-VLM} 
& \textcolor{gray}{No-CoT} & \textcolor{gray}{79.8} & \textcolor{gray}{0.0} & \textcolor{gray}{0.0} & \textcolor{gray}{24.0} & \textcolor{gray}{0.0} & \textcolor{gray}{0.0} & \textcolor{gray}{69.5} & \textcolor{gray}{0.0} & \textcolor{gray}{0.0} & \textcolor{gray}{62.3} & \textcolor{gray}{0.0} & \textcolor{gray}{0.0} & \textcolor{gray}{40.1} & \textcolor{gray}{0.0} & \textcolor{gray}{0.0} & \textcolor{gray}{82.9} & \textcolor{gray}{0.0} & \textcolor{gray}{0.0} \\
& Actual & 80.9 & 5.6 & 4.4 & 26.3 & 18.0 & 21.1 & 74.7 & 11.2 & 6.6 & 64.4 & 6.1 & 7.1 & 49.7 & 19.5 & 11.3 & 83.3 & 3.7 & 3.9 \\
& Optimal & \textbf{85.3} & \textbf{1.8} & \textbf{0.0} & \textbf{47.4} & \textbf{11.6} & \textbf{0.0} & \textbf{81.3} & \textbf{3.9} & \textbf{0.0} & \textbf{71.5} & \textbf{2.2} & \textbf{0.0} & \textbf{60.9} & \textbf{11.3} & \textbf{0.0} & \textbf{87.2} & \textbf{0.6} & \textbf{0.0} \\
\midrule
\multirow{4}{*}{MM-Eureka} 
& \textcolor{gray}{No-CoT} & \textcolor{gray}{75.8} & \textcolor{gray}{0.0} & \textcolor{gray}{0.0} & \textcolor{gray}{25.3} & \textcolor{gray}{0.0} & \textcolor{gray}{0.0} & \textcolor{gray}{67.9} & \textcolor{gray}{0.0} & \textcolor{gray}{0.0} & \textcolor{gray}{60.3} & \textcolor{gray}{0.0} & \textcolor{gray}{0.0} & \textcolor{gray}{38.6} & \textcolor{gray}{0.0} & \textcolor{gray}{0.0} & \textcolor{gray}{82.1} & \textcolor{gray}{0.0} & \textcolor{gray}{0.0} \\
& Actual & 76.4 & 19.6 & 9.6 & 32.9 & 34.0 & 13.5 & 72.8 & 20.0 & 9.6 & 64.0 & 13.1 & 7.6 & 48.2 & 11.9 & 11.3 & 82.8 & 8.6 & 5.0 \\
& Optimal & \textbf{86.0} & \textbf{6.2} & \textbf{0.0} & \textbf{46.4} & \textbf{20.1} & \textbf{0.0} & \textbf{82.4} & \textbf{7.5} & \textbf{0.0} & \textbf{71.6} & \textbf{4.6} & \textbf{0.0} & \textbf{59.5} & \textbf{6.8} & \textbf{0.0} & \textbf{87.8} & \textbf{1.5} & \textbf{0.0} \\
\midrule
\multirow{4}{*}{ThinkLite-VL} 
& \textcolor{gray}{No-CoT} & \textcolor{gray}{75.7} & \textcolor{gray}{0.0} & \textcolor{gray}{0.0} & \textcolor{gray}{21.4} & \textcolor{gray}{0.0} & \textcolor{gray}{0.0} & \textcolor{gray}{65.5} & \textcolor{gray}{0.0} & \textcolor{gray}{0.0} & \textcolor{gray}{64.1} & \textcolor{gray}{0.0} & \textcolor{gray}{0.0} & \textcolor{gray}{40.1} & \textcolor{gray}{0.0} & \textcolor{gray}{0.0} & \textcolor{gray}{82.7} & \textcolor{gray}{0.0} & \textcolor{gray}{0.0} \\
& Actual & 75.1 & 11.8 & 9.2 & 28.3 & 28.0 & 26.3 & 70.4 & 19.3 & 11.0 & 65.6 & 10.9 & 11.1 & 49.7 & 23.0 & 13.4 & 83.2 & 9.9 & 6.0 \\
& Optimal & \textbf{84.3} & \textbf{3.2} & \textbf{0.0} & \textbf{54.6} & \textbf{14.9} & \textbf{0.0} & \textbf{81.4} & \textbf{6.3} & \textbf{0.0} & \textbf{76.7} & \textbf{3.6} & \textbf{0.0} & \textbf{63.0} & \textbf{12.2} & \textbf{0.0} & \textbf{89.2} & \textbf{1.5} & \textbf{0.0} \\
\midrule
\multirow{4}{*}{VL-Rethinker} 
& \textcolor{gray}{No-CoT} & \textcolor{gray}{77.2} & \textcolor{gray}{0.0} & \textcolor{gray}{0.0} & \textcolor{gray}{28.0} & \textcolor{gray}{0.0} & \textcolor{gray}{0.0} & \textcolor{gray}{70.5} & \textcolor{gray}{0.0} & \textcolor{gray}{0.0} & \textcolor{gray}{62.7} & \textcolor{gray}{0.0} & \textcolor{gray}{0.0} & \textcolor{gray}{40.5} & \textcolor{gray}{0.0} & \textcolor{gray}{0.0} & \textcolor{gray}{82.4} & \textcolor{gray}{0.0} & \textcolor{gray}{0.0} \\
& Actual & 79.2 & 19.8 & 7.8 & 33.9 & 36.3 & 24.7 & 73.0 & 26.2 & 11.9 & 63.0 & 17.2 & 13.7 & 51.2 & 29.9 & 12.1 & 83.6 & 15.0 & 7.1 \\
& Optimal & \textbf{87.0} & \textbf{4.4} & \textbf{0.0} & \textbf{58.6} & \textbf{19.5} & \textbf{0.0} & \textbf{84.9} & \textbf{6.1} & \textbf{0.0} & \textbf{76.7} & \textbf{5.1} & \textbf{0.0} & \textbf{63.3} & \textbf{15.2} & \textbf{0.0} & \textbf{90.7} & \textbf{1.9} & \textbf{0.0} \\
\midrule
\multirow{4}{*}{R1-VL} 
& \textcolor{gray}{No-CoT} & \textcolor{gray}{70.1} & \textcolor{gray}{0.0} & \textcolor{gray}{0.0} & \textcolor{gray}{26.0} & \textcolor{gray}{0.0} & \textcolor{gray}{0.0} & \textcolor{gray}{52.9} & \textcolor{gray}{0.0} & \textcolor{gray}{0.0} & \textcolor{gray}{54.5} & \textcolor{gray}{0.0} & \textcolor{gray}{0.0} & \textcolor{gray}{26.2} & \textcolor{gray}{0.0} & \textcolor{gray}{0.0} & \textcolor{gray}{79.6} & \textcolor{gray}{0.0} & \textcolor{gray}{0.0} \\
& Actual & 71.1 & 19.8 & 8.8 & 26.6 & 45.0 & 23.4 & 62.2 & 38.9 & 14.2 & 58.5 & 15.4 & 12.3 & 52.8 & 46.4 & 15.7 & 80.3 & 11.8 & 7.8 \\
& Optimal & \textbf{79.9} & \textbf{10.3} & \textbf{0.0} & \textbf{50.0} & \textbf{25.1} & \textbf{0.0} & \textbf{76.4} & \textbf{14.1} & \textbf{0.0} & \textbf{70.8} & \textbf{5.4} & \textbf{0.0} & \textbf{68.5} & \textbf{25.0} & \textbf{0.0} & \textbf{88.1} & \textbf{2.0} & \textbf{0.0} \\

    \bottomrule
    \end{tabularx}
\end{table}

%% file: tables/early_stopping_and_efficiency.tex
\begin{wraptable}{r}{0.4\linewidth}
\vspace{-0.9em}
\centering
\small
\vspace{-5pt}
\caption{Early stopping and efficient reasoning reduce verbosity, but do not consistently reduce harmful overthinking.}
\label{tab:early_stopping_efficiency}
\footnotesize
\setlength{\tabcolsep}{4pt}
\renewcommand{\arraystretch}{1.0}
\begin{tabular}{lccc}
\toprule
\textbf{Setting} & Acc $\uparrow$ & Len $\downarrow$ & ${H}$ $\downarrow$\\
\midrule
Stopping@$\infty$ & \textbf{66.5} & 17.2 & \textbf{10.1} \\
Stopping@$5$      & 64.3 & 8.9 & 18.4  \\
Stopping@$2$      & 63.2 & \textbf{5.1} & 14.1 \\
\midrule
VL-Rethinker   & \textbf{66.6} & 23.4 & 11.1 \\
DualMind-VLM  &  66.3 & \textbf{11.2} & \textbf{8.6} \\
\bottomrule
\end{tabular}
\vspace{-1em}
\end{wraptable}

%% file: figures/stability_and_multichoice.tex
\begin{figure}[t]
\centering

\begin{minipage}[t]{0.46\linewidth}
    \centering
    \includegraphics[width=0.84\linewidth]{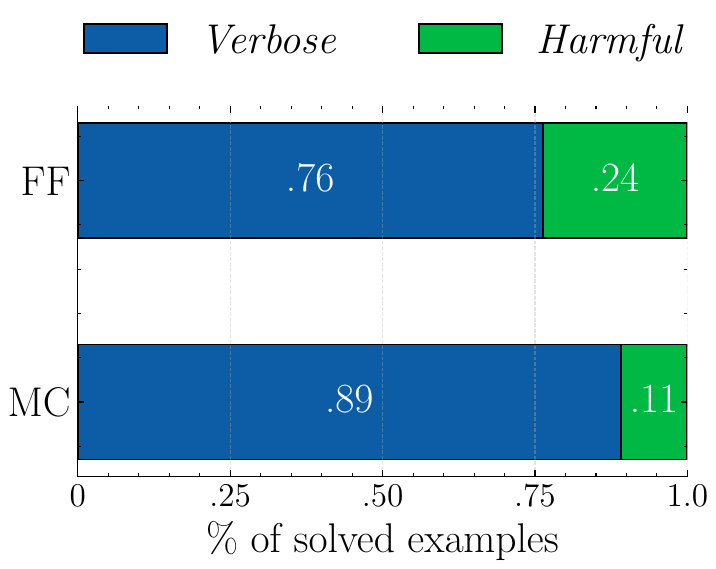}
    \captionof{figure}{Distribution of overthinking types across response formats. Bars show the percentage of solved samples exhibiting verbose versus harmful overthinking for multiple-choice (MC) and free-form (FF) settings.}
    \label{fig:mc_ff_overthinking}
\end{minipage}
\hfill
\begin{minipage}[t]{0.5\linewidth}
    \centering
    \includegraphics[width=0.84\linewidth]{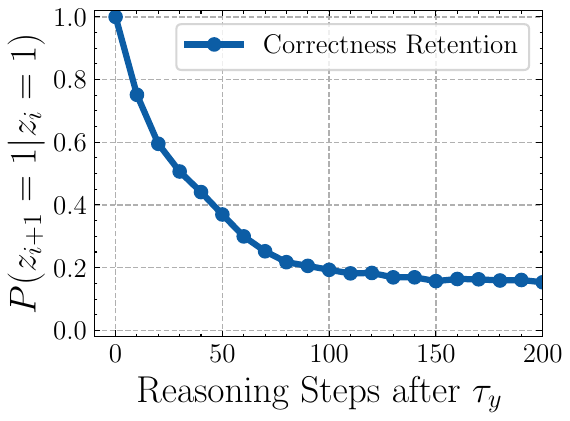}
    \captionof{figure}{Correctness stability. After first reaching a correct answer at $\tau_y$, the probability of remaining correct decreases sharply with additional reasoning, revealing diminishing reasoning value.}
    \label{fig:stability_survival}
\end{minipage}

\vspace{-1.8em}
\end{figure}

%% file: tables/difficulty_len_malign_language.tex
\begin{wraptable}{r}{0.5\linewidth}
    \centering
    \scriptsize
    \caption{
    Language-only reasoning results. We report accuracy (\(\mathrm{acc}\uparrow\)), average utterance length (\(\mathrm{len}\downarrow\)), and harmful-overthinking rate (\(H\downarrow\)).
    \emph{No-CoT} is a zero-reasoning diagnostic; bolding highlights the best nontrivial reasoning strategy. The pattern also holds for language-only models.
    }
    \vspace{2pt}
    \label{tab:main_tab_language}
    \renewcommand{\arraystretch}{1.0}
    \setlength{\tabcolsep}{3pt}
    \begin{tabularx}{\linewidth}{
        >{\raggedright\arraybackslash}p{24pt}
        >{\raggedright\arraybackslash}p{32pt}
        >{\centering\arraybackslash}X
        >{\centering\arraybackslash}X
        >{\centering\arraybackslash}X
        >{\centering\arraybackslash}X
        >{\centering\arraybackslash}X
        >{\centering\arraybackslash}X
    }
    \toprule
    \multirow{2}{*}{\textbf{Model}} 
    & \multirow{2}{*}{\textbf{Strategy}} 
    & \multicolumn{3}{c}{\textbf{GPQA}} 
    & \multicolumn{3}{c}{\textbf{AIME2025}} \\
    \cmidrule(lr){3-5} \cmidrule(lr){6-8}
    & 
    & \textit{acc$\uparrow$} 
    & \textit{len$\downarrow$} 
    & \textit{$H\downarrow$} 
    & \textit{acc$\uparrow$} 
    & \textit{len$\downarrow$} 
    & \textit{$H\downarrow$} \\
    \midrule

    \multirow{3}{*}{Qwen3} 
    & No-CoT 
    & \textcolor{gray}{37.0} & \textcolor{gray}{0.0} & \textcolor{gray}{0.0}
    & \textcolor{gray}{25.0} & \textcolor{gray}{0.0} & \textcolor{gray}{0.0} \\

    & Actual 
    & 55.8 & 125.5 & 22.1
    & 58.3 & 372.5 & 33.3 \\

    & Optimal 
    & \textbf{77.9} & \textbf{28.9} & \textbf{0.0}
    & \textbf{91.7} & \textbf{29.9} & \textbf{0.0} \\
    \midrule

    \multirow{3}{*}{InternS1} 
    & No-CoT 
    & \textcolor{gray}{37.3} & \textcolor{gray}{0.0} & \textcolor{gray}{0.0}
    & \textcolor{gray}{11.1} & \textcolor{gray}{0.0} & \textcolor{gray}{0.0} \\

    & Actual 
    & 64.4 & 177.1 & 20.3
    & \underline{38.9} & 514.2 & \underline{33.3} \\

    & Optimal 
    & \textbf{84.7} & \textbf{30.9} & \textbf{0.0}
    & \textbf{72.2} & \textbf{144.9} & \textbf{0.0} \\
    \bottomrule
    \end{tabularx}
    \vspace{-1.0em}
\end{wraptable}

%% file: sections/experiments.tex
\section{Why Does Reasoning Become Harmful?}
\label{sec:why_harmful}

The previous section shows that harmful overthinking is a systematic failure mode. \textit{But what causes a model to transition from a correct answer to an incorrect one?} In the following, we consider the multimodal setting and categorize the type of errors arising when models reason beyond the optimal.

\paragraph{Taxonomy.} For each harmful overthinking trajectory, we now identify the last correct prefix
$i^\star = \max \{ i < N : \mathcal A(t_{\le i}) = y \}$
and compare the reasoning state at $t_{\le i^\star}$ with the final trace $t_{\le N}$. 
This isolates the segment of reasoning that turns a correct trajectory into an incorrect one. 
We identify three main failure modes:
\vspace{-0.2em}

\ding{172} \noindent\emph{Visual Error.}
The model introduces an error by misreading, inventing, or over-interpreting visual evidence. 
This includes incorrect object recognition, counts, spatial relations, labels, diagram structure, or geometric interpretation.

\input{tables/taxonomy}

\vspace{-0.2em}
\ding{173} \noindent \emph{Calculation error.} The model perceives and approaches the problem correctly, but introduces an arithmetic, algebraic, unit-conversion, formula-selection, or numerical-computation error.

\vspace{-0.2em}
\ding{174} \noindent\emph{Logical error.}
The model changes its answer due to a non-visual and non-numerical reasoning failure. This includes unsupported conclusions, contradictions, irrelevant detours, answer-option mismatches, or answer revisions that are not justified by new visual or computational evidence.

\paragraph{Evaluation Protocol.}We perform this analysis on harmful overthinking cases from the same multimodal reasoning models considered in Sec.~\ref{sec:overthinking}, as well as the same benchmarks.
For each harmful trajectory, we construct a pair consisting of the last correct prefix and the final incorrect trace.
We then use an external judge model (Qwen3.6-35B) to label the dominant failure mode for each harmful trajectory and provide an \emph{evidence} of the error from the original trace.
Additional implementation details, including prompt templates and parsing rules, are provided in Appendix~\ref{supp:taxonomy_exp_det}.

\subsection{Results}

\paragraph{Quantitative.}
Table~\ref{tab:failure-modes-model-benchmark} reports the failure-mode decomposition for harmful overthinking across models and benchmarks.
Calculation errors are rarely dominant: they are never the largest failure mode for any model--benchmark pair, and often remain below \(10\%\).
Instead, harmful overthinking is primarily driven by logical drift and visual reinterpretation.
Logical errors are especially prominent on MathVerse and MathVista: on MathVerse, they exceed \(50\%\) for four out of five models and reach \(70.9\%\) for R1-VL; on MathVista, they are the leading failure mode for four out of five models.
Visual errors dominate more strongly on visually grounded benchmarks such as MathVision, MMStar, and AI2D.
For example, ThinkLite-VL reaches \(73.8\%\) visual errors on MathVision and \(64.5\%\) on AI2D, while visual errors are also the leading category for most models on MMStar.
Thus, the table suggests two recurring mechanisms behind correctness deviation: logical drift on more abstract reasoning benchmarks, and visual reinterpretation on perception-heavy ones.

\begin{takeawaybox}
\textsc{Takeaway G.} Correctness deviations are mainly driven by logical drift and visual reinterpretation rather than arithmetic mistakes.
\end{takeawaybox}

\paragraph{Qualitative.}
\input{figures/qualitatives}
We show representative examples of each failure mode in Fig.~\ref{fig:qualitatives}. In the visual-error case, the model first reaches the correct count, $\hat{y}_{i^*}=6$, but later changes its answer to $\hat{y}_{t_{\leq N}}=5$ after introducing the false observation that ``on the right side, there are 2 bricks missing.'' The subsequent arithmetic is consistent, but the visual premise is wrong.
In the calculation-error case, the model first gives the correct answer, $\hat{y}_{i^*}=65^\circ$, but later outputs $\hat{y}_{t \leq N}=61^\circ$. The added reasoning contains a direct numerical error:
$
2x = 180^\circ - 157^\circ = 23^\circ, \quad \text{so } x = 61^\circ.
$
Here the failure is not perceptual, but arithmetic introduced during the continuation.
In the logical-error case, the model first correctly answers that krill would decrease if gulls disappeared, but later changes the answer to herring. This contradicts its own explanation, which states that herring would increase. The final answer is therefore unsupported by the model's causal reasoning.
The resulting picture is that harmful overthinking is not a single error type, but different failure modes contribute to corrupting an already-correct trajectory.

%% file: tables/taxonomy.tex
\begin{table}[t]
    \centering
    \scriptsize
    \caption{
Failure-mode distribution by model and benchmark. Each triplet reports the percentage of valid harmful overthinking traces assigned to visual, calculation, or logical errors (highest in bold).
}
\vspace{1em}
    \label{tab:failure-modes-model-benchmark}
    \renewcommand{\arraystretch}{1}
    \setlength{\tabcolsep}{2pt}
    \begin{tabularx}{\linewidth}{
        >{\raggedright\arraybackslash}p{52pt}
        >{\centering\arraybackslash}X
        >{\centering\arraybackslash}X
        >{\centering\arraybackslash}X @{\hspace{10pt}}
        >{\centering\arraybackslash}X
        >{\centering\arraybackslash}X
        >{\centering\arraybackslash}X @{\hspace{10pt}}
        >{\centering\arraybackslash}X
        >{\centering\arraybackslash}X
        >{\centering\arraybackslash}X @{\hspace{10pt}}
        >{\centering\arraybackslash}X
        >{\centering\arraybackslash}X
        >{\centering\arraybackslash}X @{\hspace{10pt}}
        >{\centering\arraybackslash}X
        >{\centering\arraybackslash}X
        >{\centering\arraybackslash}X @{\hspace{10pt}}
        >{\centering\arraybackslash}X
        >{\centering\arraybackslash}X
        >{\centering\arraybackslash}X
    }
    \toprule
    \multirow{2.5}{*}{\textbf{Model}}
    & \multicolumn{3}{c}{\textbf{VMCBench}}
    & \multicolumn{3}{c}{\textbf{MathVision}}
    & \multicolumn{3}{c}{\textbf{MathVista}}
    & \multicolumn{3}{c}{\textbf{MMStar}}
    & \multicolumn{3}{c}{\textbf{MathVerse}}
    & \multicolumn{3}{c}{\textbf{AI2D}} \\
    \cmidrule(lr){2-4}
    \cmidrule(lr){5-7}
    \cmidrule(lr){8-10}
    \cmidrule(lr){11-13}
    \cmidrule(lr){14-16}
    \cmidrule(lr){17-19}
    & \textit{V} & \textit{C} & \textit{L}
    & \textit{V} & \textit{C} & \textit{L}
    & \textit{V} & \textit{C} & \textit{L}
    & \textit{V} & \textit{C} & \textit{L}
    & \textit{V} & \textit{C} & \textit{L}
    & \textit{V} & \textit{C} & \textit{L} \\
    \midrule
    DualMind-VLM
    & \textbf{50.0} & 16.7 & 33.3
    & \textbf{45.6} & 14.0 & 40.4
    & 38.7 & 16.1 & \textbf{45.2}
    & \textbf{46.5} & 9.9 & 43.6
    & 27.9 & 20.6 & \textbf{51.5}
    & \textbf{53.3} & 1.7 & 45.0 \\

    MM-Eureka
    & 45.0 & 7.5 & \textbf{47.5}
    & \textbf{50.0} & 10.5 & 39.5
    & 45.6 & 7.6 & \textbf{46.8}
    & \textbf{46.8} & 9.0 & 44.1
    & 28.4 & 14.3 & \textbf{57.3}
    & 49.0 & 0.6 & \textbf{50.3} \\

    ThinkLite-VL
    & \textbf{47.4} & 5.3 & \textbf{47.4}
    & \textbf{73.8} & 4.8 & 21.4
    & \textbf{59.1} & 6.8 & 34.1
    & \textbf{63.8} & 2.9 & 33.3
    & 38.4 & 12.4 & \textbf{49.2}
    & \textbf{64.5} & 0.0 & 35.5 \\

    VL-Rethinker
    & \textbf{46.7} & 8.0 & 45.3
    & 41.5 & 9.2 & \textbf{49.2}
    & 39.8 & 14.2 & \textbf{46.0}
    & \textbf{50.5} & 3.5 & 46.0
    & 25.3 & 16.1 & \textbf{58.6}
    & 45.3 & 0.9 & \textbf{53.7} \\

    R1-VL
    & \textbf{46.9} & 12.5 & 40.6
    & 37.2 & 9.3 & \textbf{53.5}
    & 33.3 & 19.0 & \textbf{47.6}
    & \textbf{52.2} & 3.3 & 44.4
    & 18.0 & 11.1 & \textbf{70.9}
    & \textbf{51.7} & 0.7 & 47.7 \\
    \bottomrule
    \end{tabularx}
\end{table}

%% file: figures/qualitatives.tex
\begin{figure}
    \centering
    \includegraphics[width=0.98\linewidth]{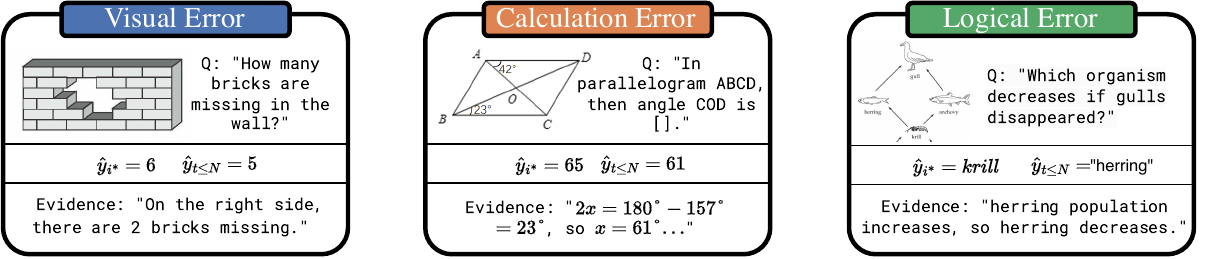}
    \caption{
Representative correctness deviations. 
Each example shows a trajectory that first reaches the correct answer at $\tau_y$, but later changes to an incorrect final answer $t_N$ through perception, calculation, or logical error. Below an evidence, representing the mistaken step of the reasoning model.
}
    \label{fig:qualitatives}
\end{figure}

%% file: sections/related.tex
\section{Related Work}

\paragraph{Test-Time Scaling and Reasoning.}
Recent reasoning models derive much of their performance from \emph{test-time scaling}: allocating more inference-time compute via longer chains of thought or larger reasoning budgets often improves accuracy \cite{muennighoff2025s1,bai2025intern,yang2025demystifying,guo2025deepseek}. 
Similar trends hold in multimodal settings, where structured deliberative traces further boost performance \cite{meng2025mmeureka,zhang2025r1,vl-rethinker,wang2025sota}. 
This line of work largely focuses on average gains from increased compute. In contrast, we study when additional reasoning is unnecessary or harmful, and when longer traces degrade rather than improve predictions.

\paragraph{Adaptive Thinking and Early Exit.}
Recent work shows that reasoning models often continue generating after reaching a correct solution, and may even revise correct intermediate states into incorrect answers~\cite{chen2024not}.
Early-exit methods stop generation using intermediate predictions, confidence, or learned signals~\cite{zhang2025reasoning,yang2025dynamic,dai2025s,fu2024efficiently}, while adaptive-thinking methods allocate variable reasoning budgets across examples using proxies such as response length or confidence~\cite{shen2025dast,zhang2025adaptthink,liu2025qfft,taubenfeld2025confidence,lin2025learning,xiao2025fastslow,wang2025make}.
Both primarily target unnecessary computation.
Our perspective is complementary: we separate \emph{verbose} overthinking, which is wasteful but harmless, from \emph{harmful} overthinking, which degrades correctness, showing that efficiency alone does not address reasoning failures.
Closest to our work, \cite{wu2025more} shows that longer CoTs do not consistently improve performance; we extend this analysis to state-of-the-art reasoning models across language and multimodal benchmarks.

\paragraph{Reasoning Compression and No-Thinking Settings.}
A related line of work questions how much explicit reasoning is required. Prior studies show that reasoning traces can often be compressed, and in some cases removed entirely without loss in performance \cite{li2026chain,ma2025reasoning,li2025think,wang2026is}. 
Our findings align with this view: the key issue is not whether models can reason longer, but whether additional reasoning is useful, redundant, or harmful.

%% file: sections/conclusion.tex
\section{Conclusion}

Test-time scaling rests on a simple premise: think longer, and performance should improve.
Our results show that this premise is incomplete, offering insights on the overlooked problem of harmful overthinking.
Across multimodal and language-only benchmarks, LRMs often reach the correct answer before termination, continue generating, and then leave the correct trajectory.
We show that optimal stopping yields large gains, many solvable instances require little or no explicit reasoning, and shorter traces fail to reduce harmful transitions.
Failure analysis shows that these errors rarely stem from arithmetic limitations; they more often arise from logical drift or visual reinterpretation.
We believe these experimental results can stimulate future work on LRMs, focusing not only on making models reason more, but also on helping them \emph{understand when reasoning is sufficient}.

%% file: sections/supplementary.tex
\section*{Supplementary Material Overview}

This appendix is organized in four macro blocks complementing the discussion in the main paper. First, Appendix~\ref{supp:robustness} provides robustness analyses for the proposed difficulty-based reasoning budget, testing the sensitivity of the first correct index estimation to sampling seeds, termination prompts, and answer extraction models. 
Second, in Appendix~\ref{supp:additional-analysis} we provide additional quantitative results, including token-level budget statistics, verbose overthinking analysis, and language-only evaluations. Third, Appendix~\ref{supp:additional-details} reports implementation and reproducibility details, such as prompt templates, parsing rules, compute accounting, and the failure-analysis categorization. Finally, Appendix~\ref{supp:limitations_future_work} discusses the limitations and possible future work.

\section{Robustness Study for Difficulty-Based Reasoning Budgets}
\label{supp:robustness}
The prefix-level trajectory evaluation protocol estimates an example's difficulty by identifying the earliest point in a model's reasoning trace from which the correct answer can be recovered. In this section, we test whether that estimate is robust to procedural choices. In particular, we measure sensitivity to three factors: the sampling seed used to generate the trace, the termination prompt used for prefix-level probing, and the answer-extraction model used to parse the answer.

\paragraph{Model and benchmark.}
We run the robustness study with VL-Rethinker on MathVision. For each condition, the model first generates a full reasoning trace for every benchmark example. We use three random seeds to measure sensitivity to stochastic generation. Raw generations are saved before answer extraction so that answer parsing can be repeated independently with different extraction models.

\paragraph{Termination prompt variants.}
\label{supp:early_termination_strategies}
The difficulty pipeline probes partial reasoning traces by appending a termination prompt that asks the model to stop deliberating and provide a final answer. We compare two variants, shown in Fig.~\ref{fig:prompt_termination_variants}: the default prompt used in our pipeline and a reworded version with the same intent and similar length. This tests whether the estimated difficulty is sensitive to a specific stop-and-answer phrase rather than reflecting the content of the reasoning trace.

\paragraph{Answer extraction variants.}
Because benchmark accuracy is computed from parsed final answers, we also vary the answer-extraction model $\mathcal{A}$. We compare \texttt{Qwen/Qwen3-4B-Instruct-2507} and \texttt{Qwen/Qwen3.5-4B} and report the answer extraction prompt in Fig.~\ref{fig:prompt_answer_extractor}. These models are used only after generation: first to parse the full CoT outputs, and later to parse the intermediate answers. This separation avoids loading the extractor during expensive VL-Rethinker generation runs and isolates parser-induced variance from reasoning-model variance.

\paragraph{Experimental design.}
The study uses three seeds, two termination prompts, and two answer extractors. For each condition, we generate raw traces, apply the corresponding answer extractor, run prefix-level difficulty probing, and evaluate correctness at each probed prefix.

\paragraph{Correlation analysis.}
To assess robustness, we compute pairwise agreement between conditions over the vector of first-correct budgets $\{b_{\tau_y}\}$ using Spearman Correlation~\cite{spearman1961proof}. High agreement across seeds indicates that the difficulty estimate is not dominated by sampling noise. High agreement across termination prompts indicates that the probing method is not overly sensitive to the exact stop-and-answer phrasing. High agreement across answer extractors indicates that the signal is not primarily an artifact of the parser. We also report the correlation of the different runs on the answer extracted at the actual length $\{z_N\}$ based on the Matthews Correlation Coefficient (MCC)~\cite{matthews1975comparison}. High agreement scores indicate that the considered conditions tend to have the same final answer.

\paragraph{Interpretation.}
The robustness results in Fig.~\ref{fig:robustness_analysis} show that the difficulty-based budget estimate is highly stable across the considered procedural variations. The Spearman correlations for the estimated optimal budget remain consistently high across all comparison groups, indicating that examples are ranked similarly by difficulty even when changing the seed, termination prompt, or answer extractor. Varying only the random seed yields high agreement, while changing the termination prompt introduces the largest drop.

A similarly stable pattern is observed for final-answer correctness at the actual reasoning length. MCC values remain close to one across all conditions, meaning that the same examples tend to be classified as correct or incorrect at the end of the full trace. The slightly lower agreement when both the answer extractor and the termination prompt change indicates that final correctness is more sensitive to parser choice and termination wording, but the effect remains small overall.

Overall, these results support the reliability of the prefix-level trajectory protocol. The estimated first-correct budgets are not artifacts of a particular sampling seed, stop-and-answer prompt, or extraction model. Instead, the high correlations suggest that the measured reasoning sufficiency signal is largely tied to the underlying reasoning trajectory.

\input{figures/robustness}

\section{Additional Analysis}
\label{supp:additional-analysis}
We provide additional analyses that complement the main results and further characterize harmful overthinking and the minimum reasoning budget for a model to answer a question.

\subsection{Correlation Between Optimal Length and No-CoT Among Models}
\label{supp:corr_opt_no_cot}

\input{figures/optimal_and_no_thinking_correlation}

Fig.~\ref{fig:optimal_and_no_thinking_corr} studies whether estimated reasoning requirements are consistent across different LRMs. Spearman's correlation on \emph{Optimal Length} is moderately high. This suggests that, despite model-specific differences in how long models reason, they often agree on the number of reasoning steps required to solve a problem.
This supports our central claim that reasoning length is a poor proxy for benchmark difficulty: many examples that elicit long traces are nevertheless perceived by several models as solvable with little or no explicit reasoning.

\subsection{Optimal Length vs.\ Test-Time Scaling}
\label{supp:test_time_scaling}

\input{figures/oracle_scaling_and_transition_minipage}

Fig.~\ref{fig:oracle_scaling} contrasts \emph{Optimal Length} with conventional test-time scaling. The test-time scaling curve, represented by \emph{Actual Length}, improves as additional samples or longer computations are allocated, but remains below the oracle \emph{Optimal Length} strategy, which stops each trajectory at its first correct prefix. This comparison shows that the limitation is not only whether the model can produce the correct answer at some point, but also whether it can preserve that answer until termination. \emph{Pass@K} provides an intermediate diagnostic: the correct answer is often present in the trajectory before the considered average length, but not always at the final utterance, corroborating the findings in Sec.~\ref{sec:overthinking}.

\subsection{Transition Matrix of Trajectories}
\label{supp:transition_matrix}

The transition matrix in Fig.~\ref{fig:transition_matrix} highlights the non-monotonic nature of reasoning trajectories moving from $t_{\leq i}$ to $t_{\leq i+1}$. If correctness were absorbing, then once a prefix was correct, later prefixes would almost always remain correct. Instead, a non-trivial number of trajectories transition from correct to wrong, showing that additional reasoning can mislead a correct intermediate solution. This is precisely the harmful-overthinking phenomenon studied in the main paper. The matrix also shows that trajectories are more likely to remain wrong than correct, further emphasizing the instability of reasoning once models leave the correct state.

\subsection{Utterances and Tokens}
\label{supp:utterances_and_tokens}

\input{figures/token_statistics}

Our main analysis uses utterances rather than raw tokens as the unit of reasoning budget. An utterance is a semantically coherent logical step in the generated trace, obtained by splitting the trace along explicit line-break delimiters (``\texttt{\textbackslash n\textbackslash n}'' and \texttt{\textbackslash n}) that LRMs naturally use when producing multi-step reasoning. This choice makes the budget less sensitive to formatting artifacts, local verbosity, and tokenizer-specific conventions. For example, two models may express the same intermediate step with different numbers of tokens, while both still represent a single reasoning transition in the trajectory.

Fig.~\ref{fig:token_statistics} reports the relationship between utterance-level and token-level budgets. The left panel shows the distribution of tokens per utterance. Most utterances are short, but the distribution has a long tail, indicating that token count can be strongly affected by unusually verbose individual steps. The right panel compares token budgets under actual length and optimal length. The same qualitative pattern observed with utterances also appears at the token level: actual traces allocate substantially more computation than is required to first reach the correct answer. Thus, our conclusions are not an artifact of measuring compute in utterances. Utterances provide a cleaner trajectory step abstraction, while token statistics confirm that the gap between actual and sufficient reasoning remains visible under a lower-level compute measure.

\subsection{On Verbose Overthinking}
\label{supp:verbose-overhead}

The main paper separates harmful overthinking from verbose overthinking. Harmful overthinking concerns correctness loss: the model reaches a correct prefix but terminates with an incorrect answer. Verbose overthinking concerns wasted computation: the model has already reached a correct answer and continues reasoning without changing the final outcome. In this Section we quantify the latter.

For each trajectory that reaches a correct prefix, we define the wasted budget as the number of utterances generated after the first correct prefix:
\[
    w(x;F) = N - \tau_y(x;F),
\]
where $N$ is the actual trace length and $\tau_y(x;F)$ is the first correct prefix. Large values indicate that the model solved the problem early but continued to spend inference compute.

Fig.~\ref{fig:wasted_budget_multimodal} reports average wasted budget across multimodal benchmarks and models. The figure shows substantial variation across models. DualMind-VLM, which is trained to decide whether to reason fast or slow, exhibits the smallest wasted budget, averaging roughly 5 unnecessary utterances. In contrast, R1-VL produces the largest wasted budget, averaging roughly 18 unnecessary utterances. However, a lower wasted budget should not be interpreted as eliminating harmful overthinking: as shown in the main results, models with shorter traces can still deviate from correct trajectories. %

\subsection{Overthinking in Language Reasoning Models}
\label{supp:language_reasoning}

The main paper shows that harmful overthinking is not restricted to multimodal reasoning. Fig.~\ref{fig:optimal_vs_actual_language} visualizes the same effect for language-only models by comparing actual and optimal utterance lengths on language benchmarks. Actual traces are extremely long, especially on mathematical reasoning tasks, whereas optimal prefixes are much shorter. This mirrors the multimodal setting: models often reach a correct solution far before their natural stopping point.

Fig.~\ref{fig:multi_choice_language} reports harmful overthinking by answer format for language-only benchmarks. The effect is again stronger in free-form settings than in multiple-choice settings. This is consistent with the multimodal results: when the output space is unconstrained, the model must preserve and express the correct answer throughout the remainder of the trace, making it more vulnerable to later revisions and contradictions.

\input{figures/wasted_budget_multimodal}
\input{figures/optimal_vs_actual_language}

\section{Additional Details}
\label{supp:additional-details}

Here, we provide the procedural details needed to reproduce our prefix-level evaluation and failure analysis. We describe the prefix-level probing setup, the taxonomy for harmful-overthinking cases, and the implementation details.

\subsection{Prefix-Level Evaluation}
\label{supp:prefix_level_evaluation}
Algorithm~\ref{alg:difficulty} summarizes the prefix-level trajectory protocol to estimate the difficulty of a sample for a given model. For each input, the model first generates a full reasoning trace. The trace is then split into utterances, and every prefix, including the empty prefix, is evaluated by appending a termination template and extracting a final answer. The returned difficulty is the first utterance index that yields a correct answer. If no prefix yields the correct answer, the instance is treated as unsolved for that trajectory.

\begin{figure}[h!]
\centering
\begin{tcolorbox}[
  promptbase,
  title={Early termination prompts}
]
\begin{lstlisting}[style=promptstyle]
P1 = "Oh, I suddenly got the answer to the whole problem.
<answer> ### **Final Answer**: \[ boxed{"

P2 = "I got it now. I can now give the final response.
<answer> ### ***Final Answer**: \[ boxed{"
\end{lstlisting}
\end{tcolorbox}
\caption{Termination prompts used for prefix-level probing. Each prompt is appended to a partial reasoning trace to force the model to stop deliberating and produce a final answer. The two variants preserve the same function while changing surface wording, allowing us to test whether estimated difficulty is sensitive to the exact probing phrase.}
\label{fig:prompt_termination_variants}
\end{figure}
\begin{figure}[t!]
\centering
\begin{tcolorbox}[
  promptbase,
  title={Answer extractor $\mathcal{A}$ prompt}
]
\begin{lstlisting}[style=promptstyle]
SYSTEM: You are a helpful assistant who extracts concise
answers from text. Extract only the direct answer provided
by the model, removing explanations.

USER: Given the following reasoning trace, extract ONLY
the final answer in a concise format.

Model Answer: {model_trace}

Extract the answer (just the answer itself, no explanations):
\end{lstlisting}
\end{tcolorbox}
\caption{Prompt used by the answer extractor $\mathcal{A}$. The variable \texttt{model\_trace} denotes the raw generation produced by the evaluated model, either at full length or after prefix-level probing. The extractor returns only the concise final answer used for benchmark verification.}
\label{fig:prompt_answer_extractor}
\end{figure}

\subsection{Taxonomy Experiment Details}
\label{supp:taxonomy_exp_det}

The taxonomy experiment analyzes harmful-overthinking cases, i.e., trajectories that reach a correct answer at some prefix but terminate with an incorrect final prediction. For each case, we identify the last correct prefix and compare it with the full final trace, thereby isolating the additional reasoning segment responsible for the correctness deviation. Fig.~\ref{fig:prompt_taxonomy_judge} summarizes the prompt configuration used to extract the category and supporting evidence.

We classify each harmful trajectory into one dominant failure mode: visual hallucination/perception error, calculation error and Logical error.

We use an external judge model, \texttt{Qwen3.6-35B}, to assign the label. The judge receives the last correct prefix, the final trace, the ground-truth metadata, and, when available, the image associated with the example. The prompt instructs the judge to compare only the reasoning added after the last correct prefix and to ignore the standardized forced-answer suffix used by the probing pipeline. The judge returns a compact JSON object containing the primary category, optional secondary categories, severity, a short explanation, evidence, and confidence. We parse only valid JSON outputs; malformed outputs are discarded or re-run under the same prompt configuration.

\input{figures/pseudocode}

\label{supp:reproducibility}

\subsection{Implementation Details and Reproducibility}
\label{app:implementation}
\paragraph{Evaluation pipeline} We re-implement and re-run all benchmark evaluations from scratch using a unified LLM-based answer-extraction pipeline. Instead of relying on benchmark-specific regular expressions, we apply a fixed answer extractor $\mathcal{A}$ to each generated trace and use the extracted concise answer for verification. This design is important because reasoning models frequently deviate from requested answer formats, and prefix-level probing produces partial traces whose answers can appear in heterogeneous forms. Unless otherwise specified, we use \texttt{Qwen/Qwen3-4B-Instruct-2507} as the extractor. Appendix~\ref{supp:robustness} repeats the difficulty-estimation analysis with \texttt{Qwen/Qwen3.5-4B} to measure sensitivity to the parser. The extractor prompt is shown in Fig.~\ref{fig:prompt_answer_extractor}.

\paragraph{Hyperparameters and Answer Extraction.} For each evaluated reasoning model, we use the reference decoding configuration recommended by the corresponding model release whenever available, including temperature, top-$p$, maximum generation length, and image-processing settings. \textit{Actual Length} denotes the model's natural termination behavior under this configuration. For prefix-level difficulty estimation, we first generate the complete reasoning trace, split it into utterances, and probe nested prefixes by appending a fixed termination template that asks the model to stop and provide a final answer. 
The failure-mode taxonomy in Appendix~\ref{supp:taxonomy_exp_det} is produced by a separate judge model, which compares the last correct prefix with the final incorrect trace and labels the newly introduced error as visual, calculation, or logical. The judge prompt explicitly instructs the model to ignore the artificial termination suffix used by the probing pipeline.

\paragraph{Compute.} All experiments are run with \texttt{vLLM} for batched inference on machines equipped with four NVIDIA A100-64GB GPUs. 
We store raw generations before answer extraction, which allows parsing, verification, robustness checks, and failure analyses to be repeated without regenerating expensive model traces. We release the evaluation scripts, prompts, decoding configurations, intermediate generations, parsed predictions, and analysis code required to reproduce the reported results.
On average an evaluation on a benchmark can span from 1 to 4 hours depending on the model and dataset size (around 1K samples on average in our setting).

\paragraph{Packages, versions, and licenses.} Our implementation was developed in Python 3.10.19, distributed under the Python Software Foundation License Version 2. We used vLLM v0.20.0 for efficient large language model inference, released under the Apache License 2.0; PyTorch v2.11.0+cu130 for tensor operations and GPU-accelerated model execution, released under a BSD-style license; and Hugging Face Transformers v5.6.2 for model and tokenizer interfaces, released under the Apache License 2.0.

\section{Limitations and Future Work}
\label{supp:limitations_future_work}

\paragraph{Verifiable outputs.}
Our analysis is limited to settings where correctness can be automatically verified, which is necessary for estimating the first correct prefix and separating verbose from harmful overthinking. The conclusions are therefore strongest for benchmarks with well-defined ground-truth answers, such as mathematical reasoning, visual reasoning, and scientific QA. Open-ended generation, tool use, and coding tasks may require different definitions of correctness. 
For example, a program can be partially correct, fail hidden tests, or improve through later debugging. 
Extension to execution-based or subjective evaluation settings is an important direction for future work.

\paragraph{Model-dependent difficulty.}
The difficulty we estimate is not an intrinsic property of a problem alone, but a property of how a particular model processes that problem. We view this model dependence as a feature of the formulation rather than only a limitation. The same problem may be easy for one model and difficult for another, depending on the model's training data, post-training procedure, visual grounding ability, mathematical knowledge, reasoning shortcuts, and decoding policy. Accordingly, the empirical difficulty $\hat{\kappa}(x,y;F)$ should be interpreted as a model-conditioned quantity: it measures the minimum reasoning budget required by model $F$, on a sampled trajectory, to recover the correct answer. This is precisely the notion we aim to capture, since overthinking is also a property of a model's own reasoning dynamics rather than of the benchmark instance in isolation.

\paragraph{Compute accounting.}
The experiments require substantial inference compute because prefix-level probing evaluates many nested prefixes for each generated trace. We report the hardware used in Appendix~\ref{supp:reproducibility} and save intermediate generations to avoid unnecessary regeneration.

\paragraph{Oracle stopping and deployability.}
\emph{Optimal Length} is not a deployable inference method because it requires ground-truth access to identify the first correct prefix. It serves as an oracle measuring how much performance is lost when models continue reasoning after a correct answer has already become recoverable. Developing practical stopping policies that approximate these oracles without access to ground truth remains an important open problem. Future work will explore how to best leverage the empirical difficulty estimate $\hat{\kappa}(x,y;F)$. It provides a possible supervision signal: models could be rewarded for reaching correct answers with sufficient but non-redundant reasoning, rather than for producing longer traces. This could support explicit stopping policies, model-agnostic difficulty predictors, or training objectives that penalize reasoning beyond the first correct prefix. The taxonomy analysis also suggests targeted interventions: visual errors may require stronger grounding, calculation errors may benefit from symbolic verification, and logical drift may require consistency constraints that prevent unsupported answer revisions.

\begin{figure}[H]
\centering
\begin{tcolorbox}[
  promptbase,
  title={Failure-analysis judge prompt configuration}
]
\begin{lstlisting}[style=promptstyle]
TAXONOMY = [
    "visual_hallucination_or_perception",
    "calculation_error",
    "logical_error",
]

PROBE_SUFFIX_INSTRUCTION = """Important probe artifact:
- Ignore forced final-answer probe suffixes that start like
  "Oh, I suddenly/finally got the answer..." and lead into "\\boxed{".
- Treat that text as evaluator scaffolding, not as reasoning produced by
  the model under test.
- Do not classify a sample as a logical error only because this standard
  probe suffix appears.
- Classify the drift using the substantive reasoning or final-answer
  change before/around that scaffold."""

COMPACT_OUTPUT_INSTRUCTION = """Output style:
- Return exactly one compact JSON object.
- No analysis, markdown, prose, or preamble.
- Use the metadata as ground truth for last/final predictions.
- Keep went_wrong to one short sentence.
- Use example for one minimal quote/paraphrase from this sample."""

PROMPT = """
You are analyzing overthinking in a nested difficulty reasoning trace.

The first trace is the LAST prefix where the model's parsed answer was
still correct. The second trace is the FINAL prefix with all retained
utterances.

Task:
1. Compare only what changed after the last-correct prefix.
2. Identify the main failure mode introduced by the final/full trace.
3. If an image is provided, decide whether the added suffix hallucinates
   or misreads visual evidence.
4. Choose the best available category even when the drift is ambiguous.
5. Ignore the standard forced final-answer probe suffix.

Allowed categories: {categories}

Return only valid JSON:
{
  "category": "one_allowed_category",
  "secondary_categories": ["zero_or_more_allowed_categories"],
  "severity": 0_to_100_integer,
  "went_wrong": "short explanation",
  "evidence": "short quote or paraphrase from the added/final trace",
  "example": "minimal quote or paraphrase illustrating the reason",
  "confidence": 0.0
}
"""
\end{lstlisting}
\end{tcolorbox}
\caption{Failure-analysis judge prompt. The judge compares the final incorrect trace against the last correct prefix and labels the dominant failure mode introduced by the additional reasoning. The prompt explicitly instructs the judge to ignore the standardized forced-answer suffix used by the probing pipeline.}
\label{fig:prompt_taxonomy_judge}
\end{figure}

%% file: figures/robustness.tex
\begin{figure}
    \centering
    \includegraphics[width=1\linewidth]{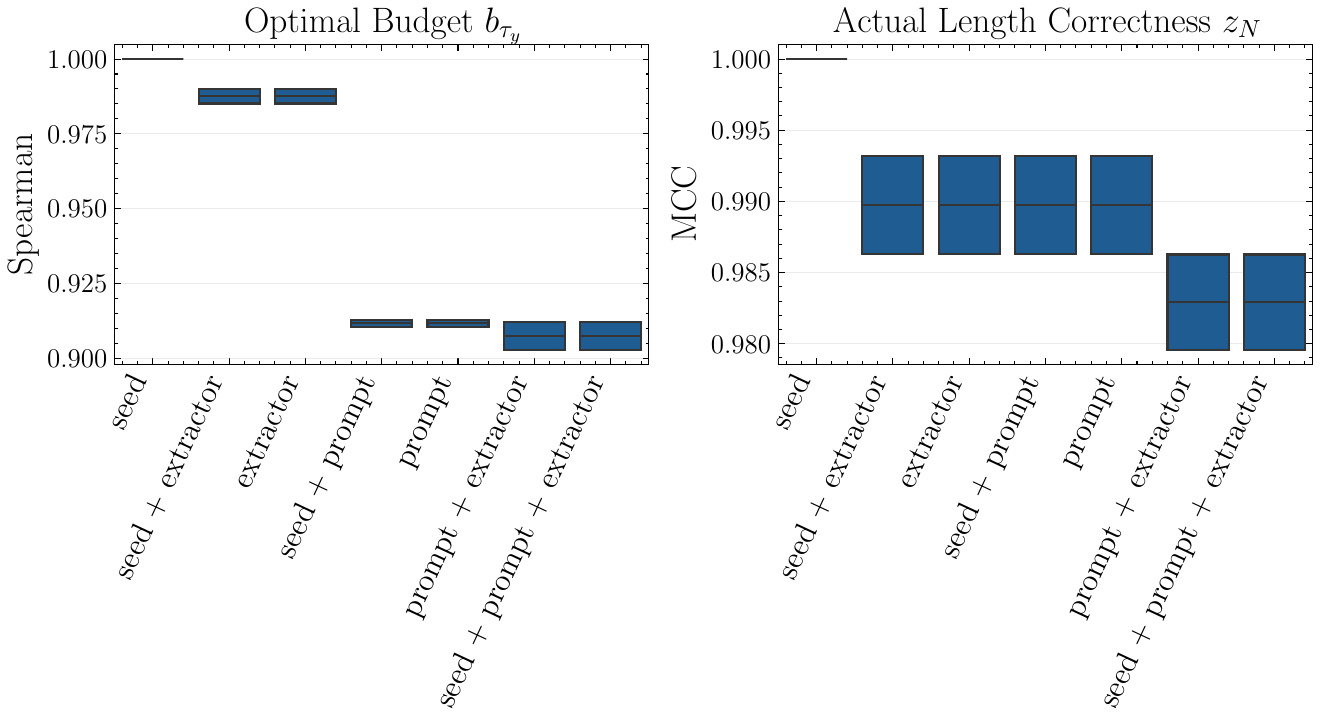}
    \caption{Robustness of the difficulty analysis across controlled sources of variation.
The left panel reports Spearman correlation of the estimated optimal budget $b_{\tau_y}$. 
The right panel reports the Matthews Correlation Coefficient for correctness of the final predicted answer $z_{N}$. 
High optimal-budget correlations indicate that examples are ranked similarly by difficulty across conditions, while high final-correctness MCC indicates that the same examples tend to be correct or incorrect at the actual reasoning length. 
The comparison groups show impact of joint variation of procedural factors.}
    \label{fig:robustness_analysis}
\end{figure}

%% file: figures/optimal_and_no_thinking_correlation.tex
\begin{figure}
    \centering
    \includegraphics[width=0.6\linewidth]{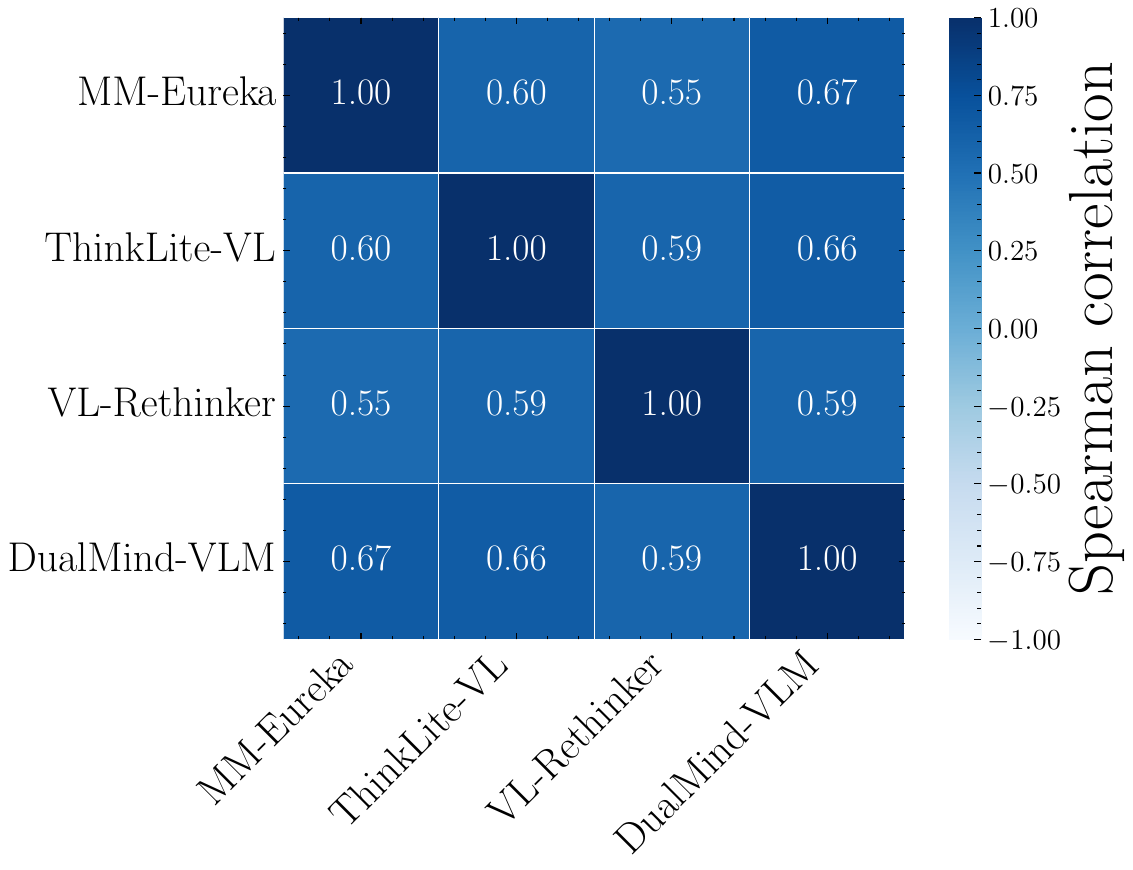}
    \caption{Cross-model Spearman Correlation in estimated \emph{Optimal Length}. Correlation on the exact optimal length is moderately high, indicating that different LRMs share a notion problem difficulty. 
    }
    \label{fig:optimal_and_no_thinking_corr}
\end{figure}

%% file: figures/oracle_scaling_and_transition_minipage.tex
\begin{figure}[t]
    \centering

    \begin{minipage}[t]{0.48\linewidth}
        \centering
        \includegraphics[width=\linewidth]{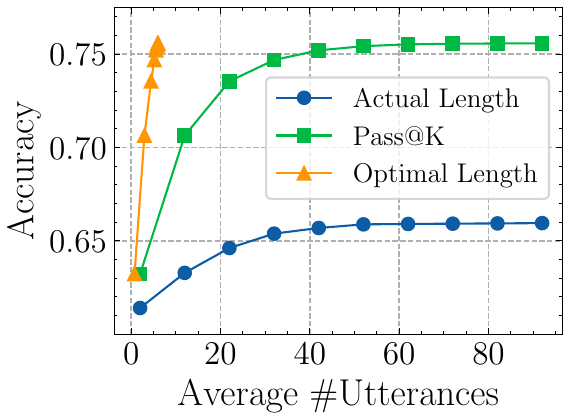}
        \caption{\emph{Optimal Length} scaling compared with standard test-time (\emph{Actual Length}) scaling and \emph{Pass@K}. Increasing test-time compute improves performance, but remains below Optimal Length, which stops each trajectory at the first correct prefix. The gap shows that models often already contain the correct answer before termination, but fail to stop before later reasoning deviates from correctness.}
        \label{fig:oracle_scaling}
    \end{minipage}
    \hfill
    \begin{minipage}[t]{0.43\linewidth}
        \centering
        \includegraphics[width=0.82\linewidth]{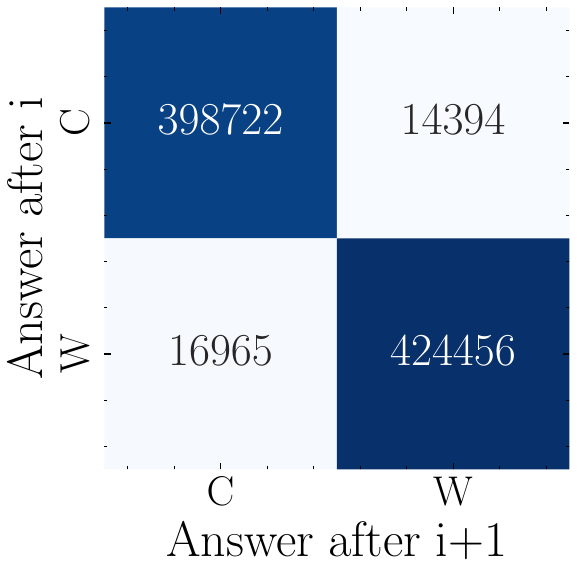}
        \caption{Prefix-level correctness transitions. Rows indicate whether the answer after prefix $i$ is correct or wrong, and columns indicate the answer after prefix $i+1$. The off-diagonal correct-to-wrong mass measures correctness deviations, showing that reasoning is not monotonic once a model has reached the correct answer.}
        \label{fig:transition_matrix}
    \end{minipage}

\end{figure}

%% file: figures/token_statistics.tex
\begin{figure}[t]
    \centering
    \begin{subfigure}{0.48\textwidth}
        \centering
        \includegraphics[width=\linewidth]{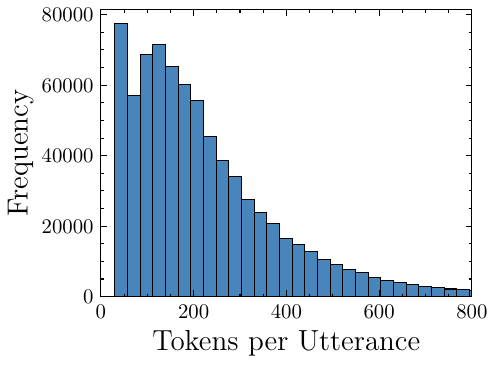}
    \end{subfigure}
    \hfill
    \begin{subfigure}{0.48\textwidth}
        \centering
        \includegraphics[width=\linewidth]{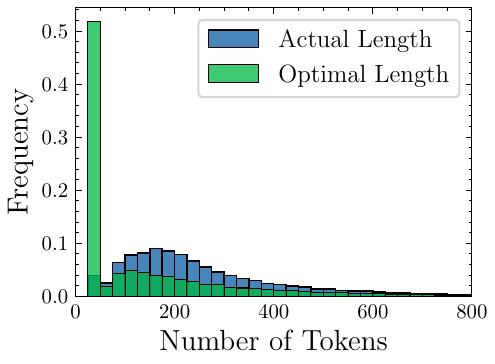}
    \end{subfigure}
    \caption{Token-level statistics for utterance-based reasoning budgets. Left: distribution of the number of tokens per utterance, showing that most reasoning steps are short but that occasional long utterances create a heavy tail. Right: token-budget distributions under \textit{Actual Length} and \textit{Optimal Length}; actual traces consume substantially more tokens than the first-correct prefixes, confirming that the utterance-level overthinking effect also appears at the token level.}
    \label{fig:token_statistics}
\end{figure}

%% file: figures/wasted_budget_multimodal.tex
\begin{figure}[t]
    \centering
    \includegraphics[width=1\linewidth]{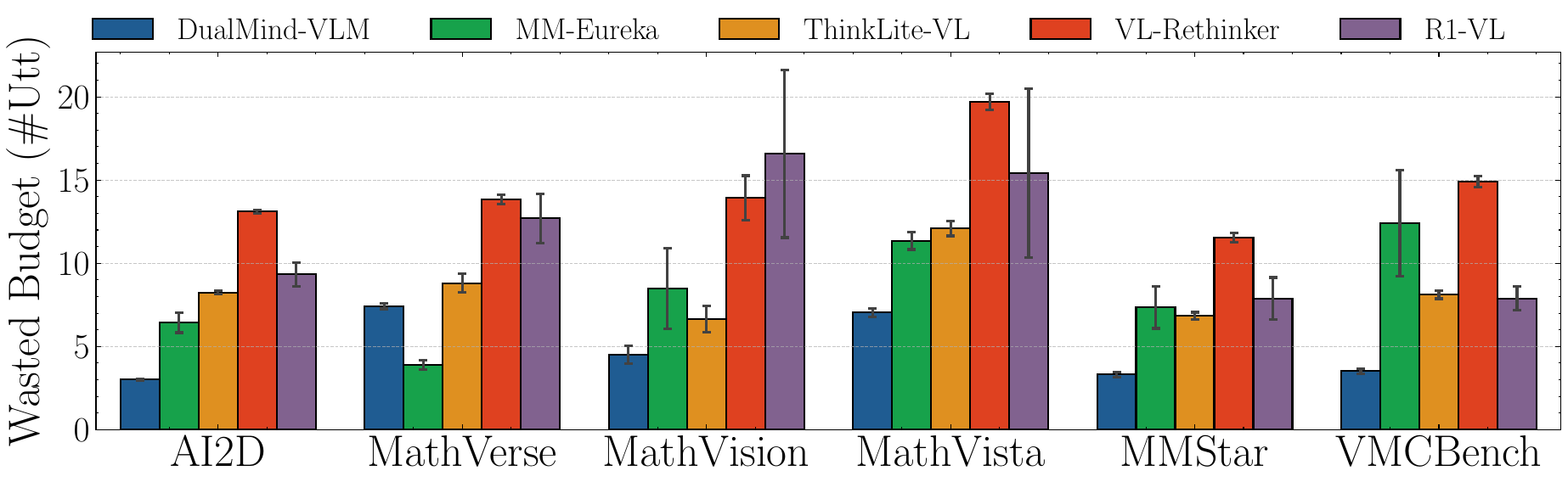}
    \caption{Average wasted budget in number of utterances per model per benchmark. DualMind-VLM~\cite{lin2025learning}, a model trained to predict input difficulty and use budget accordingly, achieves the lower wasted budget, with an average of 5 wasted utterances. R1-VL~\cite{zhang2025r1}, whose base model is Qwen2VL~\cite{wang2024qwen2}, is the least ``optimized'' model, having an average wasted budget equal to 15 utterances, while having lower base performance than all the other models~\ref{tab:main_tab}.}
    \label{fig:wasted_budget_multimodal}
\end{figure}

%% file: figures/optimal_vs_actual_language.tex
\begin{figure}[t]
    \centering

    \begin{minipage}[t]{0.45\linewidth}
        \centering
        \includegraphics[width=0.8\linewidth]{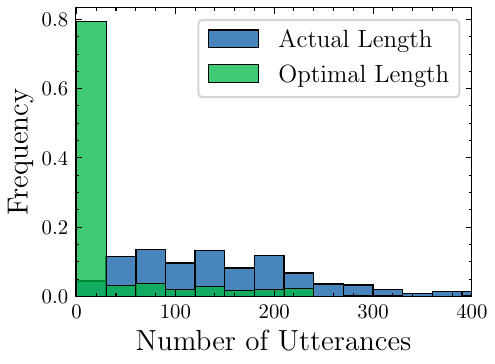}
        \caption{Actual \vs optimal reasoning length for language-only LRMs across all models and benchmarks. Actual traces are substantially longer than the first-correct prefixes, showing that language-only models also reason far beyond the point at which the correct answer first becomes recoverable.}
        \label{fig:optimal_vs_actual_language}
    \end{minipage}
    \hfill
    \begin{minipage}[t]{0.45\linewidth}
        \centering
        \includegraphics[width=0.76\linewidth]{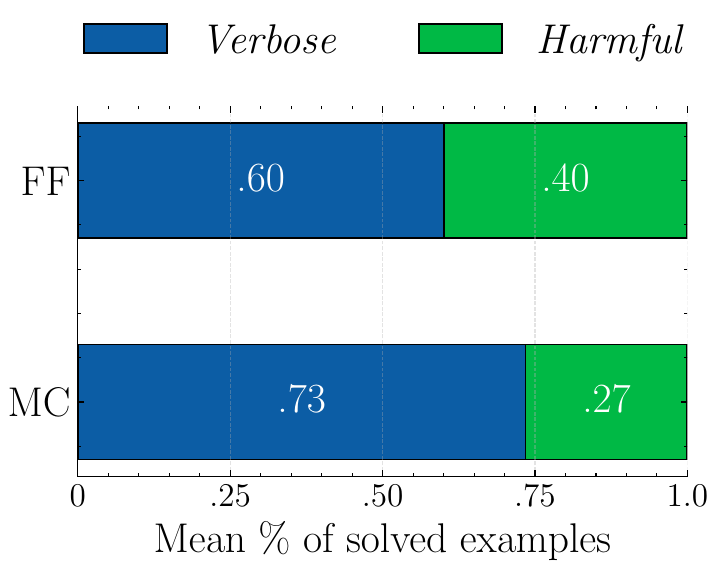}
        \caption{Harmful and verbose overthinking by answer format in language-only benchmarks. Free-form tasks exhibit higher harmful-overthinking rates than multiple-choice tasks, confirming the trend shown in the multimodal setting.}
        \label{fig:multi_choice_language}
    \end{minipage}

\end{figure}

%% file: figures/pseudocode.tex
\begin{algorithm}[t]
\caption{PyTorch-style code for $\hat{\kappa}(x; \mathcal{F})$}
\label{alg:difficulty}
\vspace{-1.ex}
\begin{lstlisting}[style=Pytorch,escapeinside={(@}{@)}]
# x = input problem
# F = reasoning model
# A = parser mapping output to a prediction
# y = ground-truth answer
# T = fixed termination template
def difficulty(x, F, A, y, T):
    # step 1: generate full reasoning trace
    t = F.generate(x)

    # step 2: split trace into utterances
    utts = split_utterances(t)

    # step 3: evaluate each prefix, including no reasoning
    for i in range(len(utts) + 1):
        prefix = "".join(utts[:i])
        prompted = prefix + T
        o_i = F.generate_from_prefix(x, prompted)
        y_hat_i = A(o_i)

        # step 4: return first correct index
        if verify(y_hat_i == y):
            return i

    # no correct prefix found
    return None
\end{lstlisting}
\vspace{-1.ex}
\end{algorithm}